\documentclass{article}

\usepackage[numbers,sort]{natbib}

    \usepackage[final]{neurips_2025}

\usepackage[utf8]{inputenc} %
\usepackage[T1]{fontenc}    %
\usepackage{hyperref}       %
\usepackage{url}            %
\usepackage{booktabs}       %
\usepackage{amsfonts}       %
\usepackage{nicefrac}       %
\usepackage{microtype}      %
\usepackage{xcolor}         %
\usepackage{float}
\usepackage{graphicx}
\usepackage{multirow}
\usepackage{multicol}
\usepackage{colortbl}
\usepackage{url}            %
\usepackage{color}
\usepackage{soul}
\usepackage{caption}
\usepackage{amsmath} 
\usepackage{array} 
\usepackage{tcolorbox}
\usepackage{makecell}
\usepackage{pifont}
\usepackage{algorithm}
\usepackage{algpseudocode}
\usepackage{placeins}
\usepackage{enumitem}
\usepackage{xspace}
\usepackage{enumitem}
\usepackage{makecell}
\usepackage{wrapfig}
\usepackage{subcaption}
\usepackage{booktabs}

\def\eg{\textit{e.g.}\@\xspace}

\algrenewcommand\algorithmicrequire{\textbf{Input:}}
\algrenewcommand\algorithmicensure{\textbf{Return:}}

\newcommand{\rowA}{\rowcolor{blue!6}}
\newcommand{\rowB}{\rowcolor{gray!20}}

\newcommand{\model}{\texttt{Struct2D}}
\newcommand{\data}{\texttt{Struct2D-Set}}

\title{Struct2D: A Perception-Guided Framework for Spatial Reasoning in MLLMs}

\author{%
  Fangrui Zhu$^{1}$\thanks{Equal contribution.}, \ Hanhui Wang$^{1, 3}$\footnotemark[1], \ Yiming Xie$^{1}$, \ Jing Gu$^{4}$, Tianye Ding$^{1}$ \\ 
 \textbf{ Jianwei Yang$^{2}$, Huaizu Jiang$^{1}$}  \\
  $^1$ Northeastern University \quad $^2$ Microsoft Research \\
  $^3$ University of Southern California $^4$ University of California, Santa Cruz \\
   \texttt{$^1$\{zhu.fang, wang.hanh, xie.yim, ding.tian, h.jiang\}@northeastern.edu}, \\
  \texttt{$^2$jw2.yang@gmail.com}, \ \texttt{$^3$hanhuiwa@usc.edu}, \ \texttt{$^4$jgu110@ucsc.edu} \\
  \vspace{-16pt}
  \href{https://github.com/neu-vi/struct2d}{\texttt{https://github.com/neu-vi/struct2d}}
}

\begin{document}

\maketitle

\begin{figure}[H]
\vspace{-2ex}
    \centering
    \includegraphics[width=\textwidth]{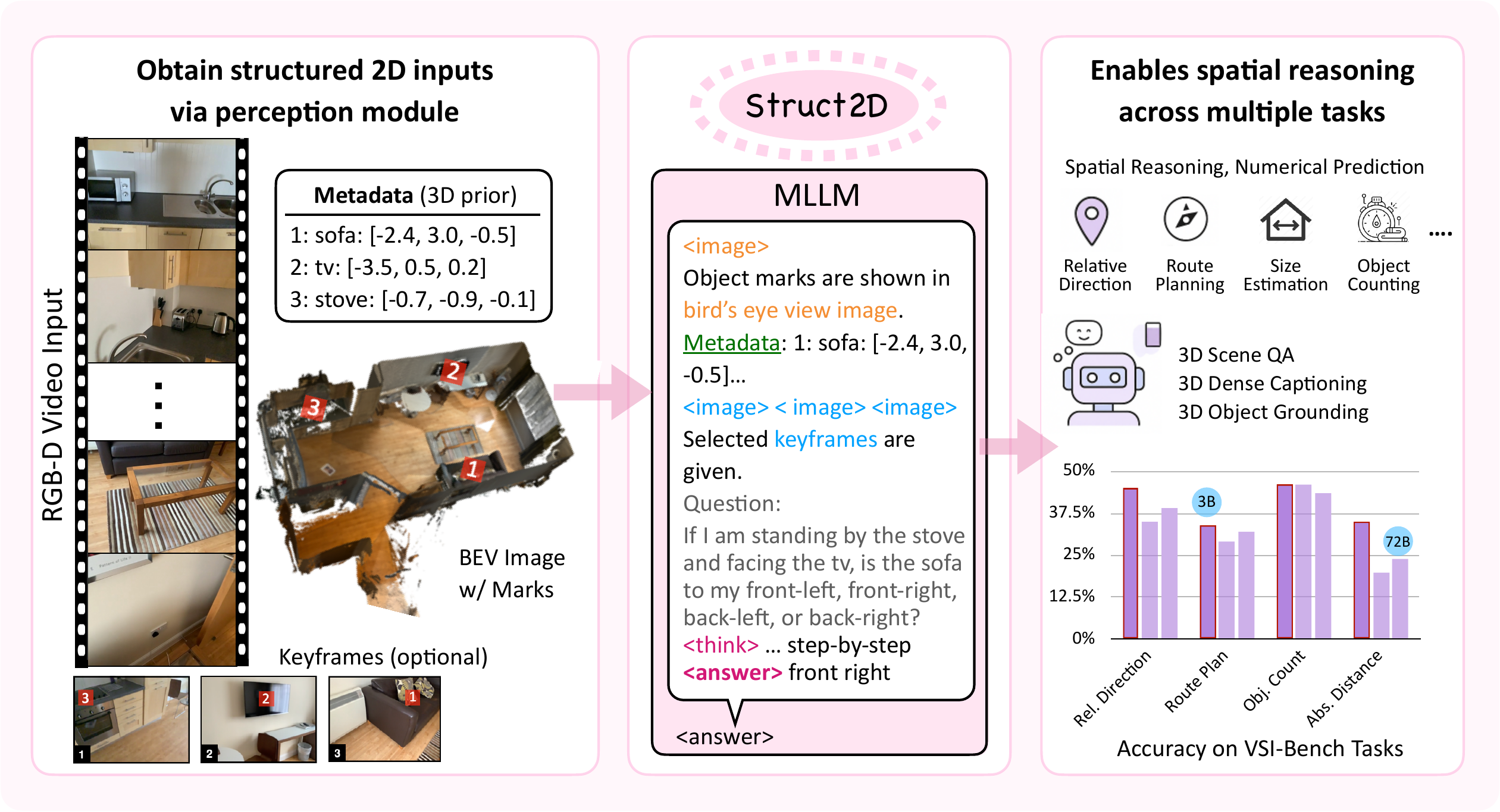}
    \caption{\label{fig:teaser} \textbf{Overview of our \model{} framework for enabling spatial reasoning in Multimodal Large Language Models (MLLMs).}
From an RGB-D video, we generate structured 2D inputs—BEV images with filtered object marks, object-centric metadata, and optional keyframes—via a 3D perception module. These inputs prompt an MLLM with spatial priors and visual context, enabling diverse spatial reasoning tasks without explicit 3D input at inference. 
   }
   \vspace{-2ex}
\end{figure}

\begin{abstract}

Unlocking spatial reasoning in Multimodal Large Language Models (MLLMs) is crucial for enabling intelligent interaction with 3D environments. While prior efforts often rely on explicit 3D inputs or specialized model architectures, we ask: \emph{can MLLMs reason about 3D space using only structured 2D representations derived from perception?}
We introduce \textbf{Struct2D}, a perception-guided prompting framework that combines bird's-eye-view (BEV) images with object marks and object-centric metadata, optionally incorporating egocentric keyframes when needed. Using Struct2D, we conduct an in-depth zero-shot analysis of closed-source MLLMs (\eg, GPT-o3) and find that they exhibit surprisingly strong spatial reasoning abilities when provided with structured 2D inputs, effectively handling tasks such as relative direction estimation and route planning.
Building on these insights, we construct \textbf{Struct2D-Set}, a large-scale instruction tuning dataset with 200K fine-grained QA pairs across eight spatial reasoning categories, generated automatically from 3D indoor scenes. We fine-tune an open-source MLLM (Qwen2.5VL) on Struct2D-Set, achieving competitive performance on multiple benchmarks, including 3D question answering, dense captioning, and object grounding.
Our approach demonstrates that structured 2D inputs can effectively bridge perception and language reasoning in MLLMs-without requiring explicit 3D representations as input. We will release both our code and dataset to support future research.

\end{abstract}
    
\section{Introduction}
\label{sec:intro}
Understanding objects and their spatial relationships in 3D space is a cornerstone of intelligent interaction in complex physical environments. Tasks such as robotic manipulation~\cite{sisbot2007spatial,landsiedel2017review}, autonomous navigation~\cite{marza2022teaching,gu2022vision}, and visual reasoning~\cite{azuma2022scanqa,yang2024thinking,ma2022sqa3d,chen2021scan2cap,chen2020scanrefer,zhu2024towards} all depend on accurate spatial understanding of scenes. At the core of these tasks lies the ability to localize objects precisely and reason about their configurations in 3D space. Moreover, grounding such spatial understanding in natural language enhances an AI system’s ability to interpret, explain, and act upon spatial information in human-centric contexts.

Traditional task-specific models rely on explicit 3D representations as input, such as point clouds or reconstructed environments~\cite{azuma2022scanqa,ma2022sqa3d,jin2023context,zhu20233d}, providing detailed geometric information. 
However, these models, often trained on limited data sources, making them less adaptable and struggle to generalize to diverse and complex textual queries. 
As a result, they fail to effectively bridge spatial reasoning with language comprehension, limiting their applicability for embodied AI.

In recent years, Multimodal Large Language Models~(MLLMs)~\cite{zhang2024llavanext-video,o3,gemini,li2024llava} developed with Large Language Models~(LLMs) have achieved significant advances in perception and reasoning tasks for images and videos. To extend MLLMs' capabilities to 3D understanding, point cloud-based LLMs~\cite{guo2023point,qi2024shapellm,qi2024gpt4point,xu2024pointllm,chen2024ll3da,fu2024scene,hong20233d,man2024lexicon3d,wang2023chat,huang2023chat} have emerged, incorporating 3D spatial features by aligning point cloud data with LLMs.
This integration enhances spatial reasoning and provides a richer understanding of the 3D physical world. However, they often rely on well-annotated datasets for instruction tuning and require point-cloud features as input, which limits their flexibility.

Unlike models that take explicit 3D representations as input, humans perceive the world as a continuous stream of 2D visual inputs akin to a \textit{video}, and naturally infer spatial relationships and object configurations by building mental representations subconsciously~\cite{nadel2008hippocampus,tolman1948cognitive}. 
Naturally, we ask ``\emph{Can MLLMs perform spatial reasoning  \emph{without} using explicit 3D features as direct inputs?}'' 
Recent work has begun to explore this direction by leveraging cognitive maps~\cite{yang2024thinking} and Bird’s Eye View~(BEV) images~\cite{qi2025gpt4scene} generated from video as 2D spatial cues, enabling MLLMs to perform spatial reasoning~\cite{yang2023set,yang2023dawn}. While promising, these approaches often omit object appearance and detailed priors (\eg, coordinates, categories), which are critical for comprehensive 3D understanding.

We conduct an in-depth analysis of MLLMs’ spatial reasoning abilities using a \textbf{perception-guided 2D framework} called \model{} Prompting. This strategy transforms 3D perception outputs—obtained from off-the-shelf detectors—into structured 2D inputs, consisting of (1) a rendered bird’s-eye-view (BEV) image with projected object marks\footnote{We follow the term ``mark'' as used in~\cite{yang2023set}.} and (2) object-centric metadata such as category labels and 3D coordinates. When appearance cues are needed, we optionally incorporate egocentric keyframes selected based on object visibility. This design enables MLLMs to reason about complex 3D scenes using only structured 2D visual and textual cues, eliminating the need for explicit 3D inputs. We begin by evaluating this approach on GPT-o3~\cite{o3}, a representative closed-source MLLM, to assess its zero-shot spatial reasoning capabilities.

To better understand the spatial reasoning capabilities of existing MLLMs, we begin with a zero-shot analysis using our proposed \model{} Prompting strategy. The goal is to evaluate whether a pretrained, closed-source model such as GPT-o3 can accurately infer 3D spatial relationships when given only structured 2D visual and textual inputs. We use rendered bird’s-eye-view~(BEV) images with projected object marks and object-centric metadata, allowing the model to reason about 3D scenes without access to explicit 3D features.
This analysis yields several key insights:
(1) A single informative BEV image, combined with metadata, is often sufficient for accurate zero-shot 3D scene understanding;
(2) Prompt composition is critical—different spatial reasoning tasks benefit from tailored input formats;
(3) For challenging tasks in VSI-Bench~\cite{yang2024thinking}, such as egocentric-to-allocentric transformations, MLLMs can perform robustly when provided with well-structured 2D projections of the 3D scene.

Guided by the findings from our zero-shot analysis, we construct a large-scale instructional tuning dataset, named \data{}, using an automated pipeline. The dataset consists of 200K QA pairs generated from 6K 3D indoor scenes, leveraging ground-truth object annotations provided by the original 3D datasets. It spans eight categories of spatial reasoning tasks relevant to embodied AI. To ensure data quality, we use ChatGPT to both enrich the QA pairs with step-by-step reasoning traces and identify potentially low-quality samples. Additionally, we incorporate a human-in-the-loop review process to further refine and validate the dataset.
We then fine-tune an open-source MLLM (Qwen2.5VL~\cite{wang2024qwen2}) using \data{}. Although the fine-tuned model is evaluated under noisy 3D perception conditions, it achieves strong performance across multiple spatial reasoning benchmarks, including 3D question answering~\cite{yang2024thinking,azuma2022scanqa,ma2022sqa3d}, spatial captioning~\cite{chen2021scan2cap}, and object grounding~\cite{chen2020scanrefer,Multi3DRefer}, demonstrating the practicality and robustness of our approach.

Our main contributions are as follows:
\begin{itemize}[leftmargin=*, nosep]
\item We propose a perception-guided 2D prompting strategy, \model{} Prompting, and conduct a detailed zero-shot analysis that reveals MLLMs’ ability to perform 3D spatial reasoning from structured 2D inputs alone.
\item We introduce \data{}, a large-scale instructional tuning dataset with automatically generated, fine-grained QA pairs covering eight spatial reasoning categories grounded in 3D scenes.
\item We fine-tune an open-source MLLM to achieve competitive performance across several spatial reasoning benchmarks, validating the real-world applicability of our framework.
\end{itemize}

\section{Related Work}
\label{sec:related_work}

\noindent \textbf{3D Spatial Reasoning with MLLMs.}
Developing real-world embodied agents requires equipping Multimodal Large Language Models (MLLMs) with robust 3D spatial reasoning abilities~\cite{cai2024spatialbot,chen2024spatialvlm,cheng2024spatialrgpt,li2024topviewrs,liu2024coarse,yang2024v,zhu2024llava}.
Recent efforts have explored spatial understanding through language~\cite{momennejad2023evaluating,rozanova2021grounding,wu2024visualization}, static 2D images~\cite{ramakrishnan2024does,tang2024sparkle,yang2023set,marsili2025visual,ma2025spatialreasoner}, or videos~\cite{qi2025gpt4scene, yang2024thinking, liao2025improved, gu2025phyworldbench}.
Our work builds upon the video-input setting, but diverges by enabling spatial reasoning in MLLMs using only structured 2D inputs—BEV images, object marks, and metadata—without relying on explicit 3D encoder / representations at inference.

\noindent \textbf{Instruction Tuning for 3D Spatial Reasoning.}
Recent work~\cite{Robin3D,li2023m3dbench,li20243dmit,chen2024ll3da,liao2025improved} has explored instruction tuning to enhance MLLMs’ capabilities for 3D spatial reasoning, targeting tasks such as 3D visual grounding~\cite{achlioptas2020referit3d,chen2020scanrefer,Multi3DRefer}, 3D dense captioning~\cite{chen2021scan2cap}, and 3D question answering~\cite{azuma2022scanqa,ma2022sqa3d}.
M3DBench~\cite{li2023m3dbench} provides region- and scene-level instruction-response pairs for general 3D understanding, while 3DMIT~\cite{li20243dmit} focuses on scene-centric instructions. LL3DA~\cite{chen2024ll3da} supports interactive planning and reasoning across omni-3D inputs. Robin3D~\cite{Robin3D} introduces a 3D LLM trained on diverse instruction-following examples.
R1-Zero-VSI~\cite{liao2025improved} proposes a video-based instruction tuning dataset and a GRPO-based training method to enhance spatial reasoning in Qwen2-VL, yet its QA pairs involve limited reasoning complexity, cover fewer task types, and yield marginal performance gains.
In contrast, we propose \data{}, a large-scale dataset that enables open-source MLLMs to acquire rich 3D spatial reasoning skills through instruction tuning—using only structured 2D representations, without requiring direct access to 3D point clouds.

\noindent \textbf{3D Point Cloud LLMs.}
Recent advances in 3D point cloud LLMs enable natural language generation and interaction grounded in 3D geometry by directly processing point clouds as input. These models benefit from the geometric precision and texture richness of point clouds, offering stronger spatial understanding than raw image or video inputs.
Prior work has focused on object-level~\cite{guo2023point,qi2024shapellm,qi2024gpt4point,xu2024pointllm} and scene-level~\cite{chen2024ll3da,fu2024scene,hong20233d,man2024lexicon3d,wang2023chat,huang2023chat} spatial reasoning.
However, directly using point cloud features requires additional training and infrastructure, limiting flexibility and scalability in real-world applications.

\noindent \textbf{Prompting LLMs.}
Despite the rapid scaling of large language models (LLMs)\cite{brown2020language,chowdhery2023palm,touvron2023llama,zhang2022opt,achiam2023gpt,hurst2024gpt,dubey2024llama}, their reasoning capabilities remain heavily dependent on effective prompt design. In-context learning\cite{brown2020language,dong2022survey}, which conditions models on a few representative examples, has become a widely adopted technique for improving instruction-following behavior.
To further enhance reasoning, strategies such as chain-of-thought~\cite{wei2022chain} and tree-of-thought~\cite{yao2023tree} prompting have been proposed.
More recently, Multimodal Large Language Models (MLLMs)\cite{o3,gemini,liu2023visual,zhu2023minigpt,zhang2023llama,dai2023instructblip,li2024multimodal,chen2024internvl,xue2024longvila,lin2023vila,zhang2024longva,zhang2024llavanext-video,li2024llava} have gained prominence for their ability to reason over multiple input modalities. This has led to a surge of research into prompting techniques tailored for MLLMs\cite{yang2023dawn,yang2023set,cheng2024spatialrgpt,cai2024vip,hu2025visual,nasiriany2024pivot,wu2025controlmllm,wu2024dettoolchain,yu2024attention,hu2025leveraging,lin2024draw,lee2024affordance,zhang2024earthmarker,zheng2024tracevla,liu20233daxiesprompts,sun2024layoutvlm,wu2024visual,yang2025magma}.
Building on this direction, we propose \model{}, a structured 2D prompting strategy that enables MLLMs to perform 3D spatial reasoning effectively—without requiring explicit 3D input representations.

\section{Analysis on \model{} prompting with GPT-o3} 
\label{sec:zero-shot}
\begin{figure}
    \centering
    \includegraphics[width=\linewidth]{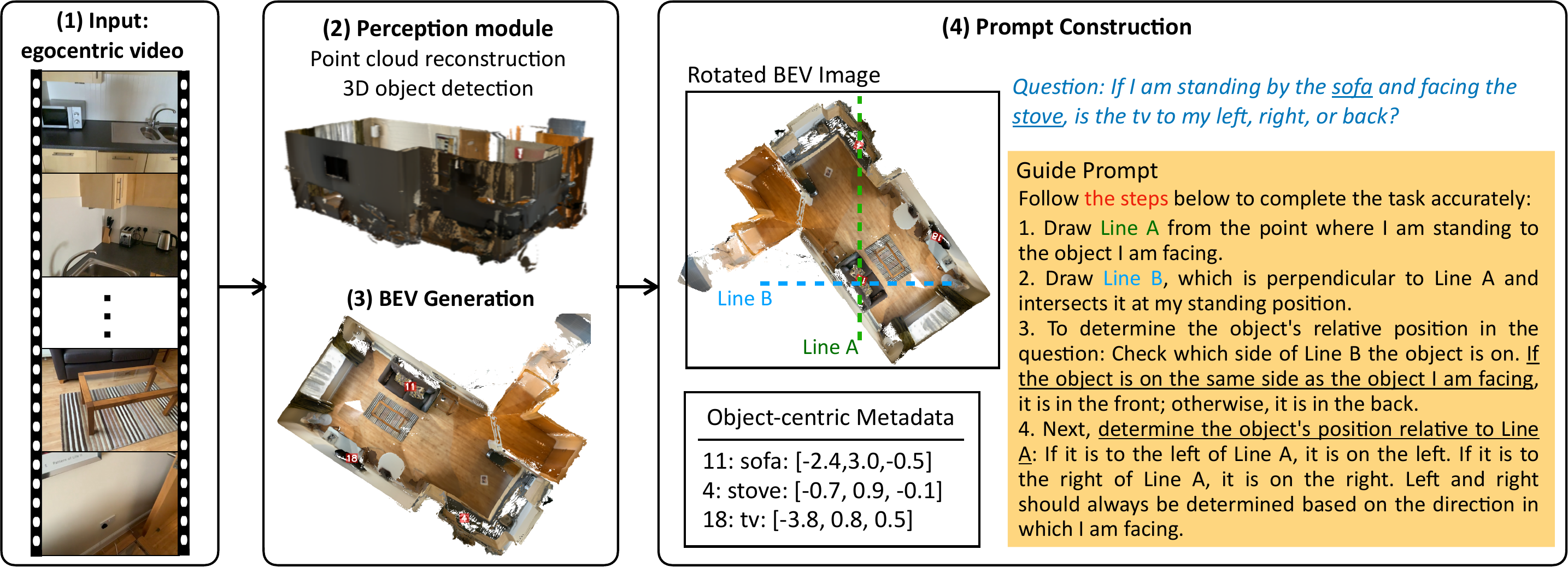}
    \caption{\label{fig:prompt}\textbf{Illustration of \model{} prompting.} Given an egocentric video, we first reconstruct a point cloud and detect 3D objects. A bird’s-eye-view (BEV) image is rendered and drawn with object marks related with the question. To facilitate reasoning about relative directions, the BEV is rotated to align with the agent’s facing direction. We further construct object-centric metadata and a structured guide prompt to support the model in understanding spatial relationships between objects. }
\end{figure}
\subsection{\model{} Prompting}
Given a video $\mathbf{V}$ as input, an MLLM $\mathcal{F}$ processes a set of $N$ sampled video frames, denoted as $\mathbf{I} = \{I_1, I_2, \dots, I_N \}$, where each frame $I_n$ has dimensions $\mathbb{R}^{H \times W \times 3}$ for $n \in \{1, ..., N\}$. Alongside visual input, the MLLM receives a text query of length $l_i$, represented as $\mathbf{T}^\text{in} = [t^i_1, \dots, t^i_{l_i}]$. The model then generates a textual response of length $l_o$, denoted as $\mathbf{T}^\text{out} = [t^o_1, \dots, t^o_{l_o}]$, formulated as:

\begin{equation} 
    \label{eq:llm}
    \mathbf{T}^\text{out} = \mathcal{F}(\mathbf{I}, \mathbf{T}^\text{in}).
\end{equation}

However, directly using video frames for spatial reasoning introduces two major limitations:
(1) \textbf{Incomplete perception} — Video frames are typically sampled sparsely and from limited viewpoints, which can result in missing critical visual evidence required for spatial understanding. For instance, consider a scene where a chair is tucked partially under a table. If most sampled frames are taken from frontal views or from a standing height, the chair's presence might be obscured or entirely invisible, leading the model to incorrectly assume there is empty space beneath the table. This limitation becomes more severe in cluttered or occluded environments, where small objects or those blocked by other furniture.
(2) \textbf{Lack of global context} — Video frames offer fragmented, egocentric views that often fail to capture the overall structure of the scene. For example, determining whether a lamp is closer to the couch or the bookshelf may be impossible if the two objects never co-occur in the same frame. Without a consistent top-down or holistic representation, the model must rely on spatial memory or reasoning across disjoint perspectives—an ability that remains weak in most MLLMs. This fragmentation also impedes the understanding of traversability (\eg, identifying a clear path from the door to the kitchen) or relational queries (\eg, which chair is directly behind the dining table).

To address these issues, \model{} incorporates a perception module $\phi_{\text{percept}}$ 
that extracts point clouds $\mathcal{P}$ and object detections $\mathcal{O}$ from the input video $\mathbf{V}$. We then generate a top-down bird's-eye-view image with filtered object marks—only including objects relevant to the question, as illustrated in Figure~\ref{fig:prompt}.
Additionally, we construct object-centric metadata $\mathbf{T}^\text{meta}$ (\eg, categories, coordinates) as textual input to guide spatial reasoning.
Formally, we redefine Eq.~\ref{eq:llm} as:
\begin{equation}
\label{eq:llm_final_1}
\mathbf{T}^\text{out} = \mathcal{F}(\text{\model}(\phi_{\text{percept}}(\mathbf{V}), \mathbf{T}^\text{meta}),\mathbf{T}^\text{in}).
\end{equation}

For questions requiring appearance or depth cues~(\textit{e.g.}, object color or size), we supplement the BEV view with selected egocentric keyframes $\mathbf{I}_{\text{keyframe}}$ that capture clear views of the relevant objects. 
Instead of uniformly sampling keyframes, we use 3D projections to select frames that better capture the spatial coverage of the scene.
The full formulation becomes:
\begin{equation}
\label{eq:llm_final_2}
\mathbf{T}^\text{out} = \mathcal{F}(\text{\model}(\phi_{\text{percept}}(\mathbf{V}), \mathbf{T}^\text{meta}, \mathbf{I}_{\text{keyframe}}),\mathbf{T}^\text{in}).
\end{equation}
This formulation illustrates how \model{} leverages 3D perception as an intermediate step to generate informative 2D inputs that preserve spatial structure. Although 3D point clouds are used during preprocessing, they are not directly provided to the MLLM. Instead, they are transformed into BEV images and metadata used for prompting. As a result, the model performs spatial reasoning effectively without requiring explicit 3D representations as input.

\begin{wrapfigure}{r}{0.45\linewidth}
    \centering
    \includegraphics[width=\linewidth]{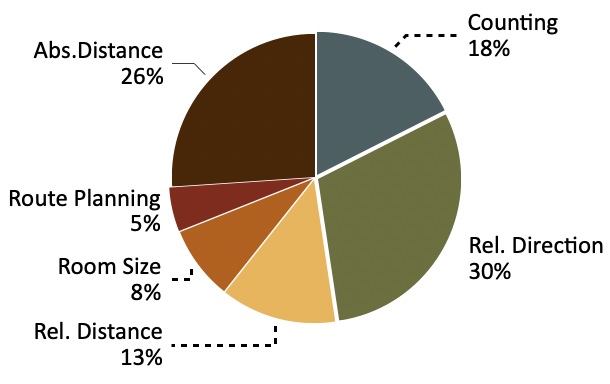}
    \caption{\label{fig:subset_dist}\textbf{Distribution of question types in the selected VSI-Bench subset.} This follows the distribution of the full set.}
    \vspace{-5pt}
\end{wrapfigure}
\textbf{Evaluation Setup.}
We sample questions from VSI-Bench~\cite{yang2024thinking}, which is designed to evaluate complex spatial reasoning skills. Compared to traditional 3D QA datasets~\cite{azuma2022scanqa,ma2022sqa3d}, VSI-Bench covers more fine-grained object perception requirements, intricate global spatial relationships, and egocentric-to-allocentric transformations. It also features diverse indoor scene sources and robust evaluation metrics that go beyond rule-based NLP scoring.

\textbf{Comparison to GPT4Scene Prompting~\cite{qi2025gpt4scene}.} While GPT4Scene pioneered 2D spatial prompting using BEV images, our \model{} strategy introduces several key improvements:
(1) \underline{Filtered object marks} tailored to the query improve visual relevance and reduce distraction;
(2) \underline{Guided metadata prompts} provide additional spatial priors;
(3) \underline{Keyframe selection} is optimized using depth-aware 3D projection instead of uniform sampling, making them both fewer and more informative~(training drops from 6 to 4 hours).

\subsection{Zero-shot Analysis of \model{} Prompting}

We construct a subset of 422 QA pairs for evaluation, selected due to API call budgets. As shown in Figure~\ref{fig:subset_dist}, the distribution of question types is consistent with the full benchmark. For our analysis, we generate object marks using both ground-truth 3D annotations and noisy detections (following~\cite{huang2023chat,qi2025gpt4scene}), ensuring comprehensive object coverage while eliminating perception errors. This also enables a fair comparison with prior work, particularly \cite{qi2025gpt4scene}.

\begin{table}
    \centering
    \caption{\label{tab:zero_shot_subset} \textbf{Zero-shot evaluation of GPT-o3 on the VSI-Bench subset.} The first row simply uses 16 frames from the input video, proposed in VSI-Bench~\cite{yang2024thinking}. For our prompting, we only input a BEV image with object marks on it along with object-centric meta information.}
    \resizebox{\linewidth}{!}{
    \begin{tabular}{l| c c c c c c c c c c c}
        \toprule
        \multirow{2}{*}{\textbf{Settings}} & \multirow{2}{*}{\textbf{\# images}} & \multirow{2}{*}{\textbf{Cost~(\$)}} & \multirow{2}{*}{\textbf{Avg.}} & \multicolumn{3}{c}{\textbf{Numerical Answer}} & \multicolumn{3}{c}{\textbf{Multiple-Choice Answer}} \\
        \cmidrule(lr){5-7} \cmidrule(lr){8-10}
        & & & &  Obj. Count & Abs. Dist. & Room Size & Rel. Dist. & Rel. Dir. & Route Plan \\
        \midrule
           VSI-Bench~\cite{yang2024thinking} & 16 & 105.07 & 48.6 & 44.3 & 34.1 & 50.9 & 51.0 & 49.4 & 61.9  \\
     GPT4Scene~\cite{qi2025gpt4scene} & 9 & 78.67  & 50.3 & 51.5 & 35.3 & \textbf{58.0} & 50.5 & 47.9 & 58.8 \\   
     \midrule
    \rowB Ours~(Noisy Objects) & 1 & \textbf{27.25} & \textbf{56.1} & \textbf{52.8} & \textbf{38.4} & 48.9 & \textbf{60.0} & \textbf{60.1} & \textbf{76.2} \\
    \rowB Ours~(GT Objects) & 1 & \textbf{27.25} & 83.8 & 93.8 & 90.6 & 47.4 & 96.5 & 94.4 & 80.1 \\
        \bottomrule
    \end{tabular}
    }  
\end{table}

\textbf{Zero-shot Prompting Results.}
Table~\ref{tab:zero_shot_subset} shows that GPT-o3 exhibits strong spatial reasoning capabilities when prompted with structured 2D inputs. Specifically, providing both object-centric metadata and filtered object marks significantly boosts performance, achieving 96.5 on relative distance, 94.4 on relative direction, and 80.1 on route planning. This highlights that explicit 3D representations are not strictly necessary—MLLMs can reason effectively with carefully structured 2D projections.
The ablation further reveals that rotation alignment and a structured guide prompt each contribute to improved accuracy on relative direction tasks, with the combination of both yielding the best performance (94.4). These results underscore the importance of aligning spatial context and guiding the model through geometric reasoning steps.
Notably, our method requires only a single BEV image and lightweight metadata, making it a low-cost and robust alternative to multi-frame prompting strategies~\cite{yang2024thinking,qi2025gpt4scene}.

\begin{table}[h]
    \centering
    \small
    \caption{\label{tab:zero-shot-ablation}\textbf{Ablation on different prompting strategies.}}
    \begin{subtable}[t]{0.48\textwidth}
        \centering
        \setlength{\tabcolsep}{5pt}
        \begin{tabular}{@{}ccccc@{}}
            \toprule
            Metadata & \makecell{Filtered\\Marks} & Rel. Dist. & Rel. Dir. & Route Plan \\
            \midrule
            -- & -- & 67.5 & 82.1 & 74.3\\
            -- & \checkmark         & 72.1  & 88.3  & 78.3 \\
            \checkmark & --   & 75.3  & 89.5  & 50.6 \\
            \checkmark & \checkmark    & \textbf{96.5}  & \textbf{94.4}  & \textbf{80.1} \\
            \bottomrule
        \end{tabular}
        \caption{\textbf{Effects of metadata and filtered marks.}}
    \end{subtable}
    \hfill
    \begin{subtable}[t]{0.48\textwidth}
        \centering
        \setlength{\tabcolsep}{5pt}
        \begin{tabular}{@{}ccc@{}}
            \toprule
            \makecell{Guide\\Prompt} & Rotation & Rel. Dir. \\
            \midrule
            -- & -- & 75.3 \\
            -- & \checkmark     & 89.2 \\
            \checkmark & --        & 80.2 \\
            \checkmark & \checkmark          & \textbf{94.4} \\
            \bottomrule
        \end{tabular}
        \caption{\textbf{Effects of rotation and guide prompt.}}
    \end{subtable}
\end{table}

\vspace{-5pt}
\textbf{What makes a good prompt for spatial reasoning?} Table~\ref{tab:zero-shot-ablation} highlights the impact of key components in our prompting strategy. 
Incorporating object-centric metadata consistently improves performance across tasks—raising relative distance accuracy from 67.5 to 96.5 and route planning from 74.3 to 80.1—highlighting its importance for grounding spatial context. Filtering object marks based on question relevance further reduces ambiguity, yielding substantial gains in route planning (from 50.6 to 80.1).
For relative direction, both the use of a structured guide prompt and rotation alignment prove essential. While each individually improves accuracy (89.2 and 80.2 respectively), their combination leads to the best performance (94.4).
We focus on these question types in ablation because they represent core challenges in spatial understanding.

\section{Large-Scale Instruction Tuning with \data{}}

Building on the insights from our zero-shot analysis (Sec.~\ref{sec:zero-shot}), we construct a large-scale instruction tuning dataset, \data{}, tailored to support diverse spatial reasoning tasks grounded in realistic 3D indoor scenes. Notably, the dataset is designed to require only 2D projected inputs during training, enabling efficient supervision without reliance on full 3D data.

In this section, we first describe the design and construction of \data{}, highlighting its coverage, annotation pipeline, and task diversity. We then present the supervised fine-tuning (SFT) setup using open-source MLLMs, detailing the model configurations and training procedures. Finally, we evaluate the effectiveness of our instruction-tuned model across multiple spatial reasoning benchmarks, assessing its generalization and reasoning capabilities.

\subsection{\data{}}
\begin{wrapfigure}{r}{0.6\linewidth}
    \centering
    \includegraphics[width=\linewidth]{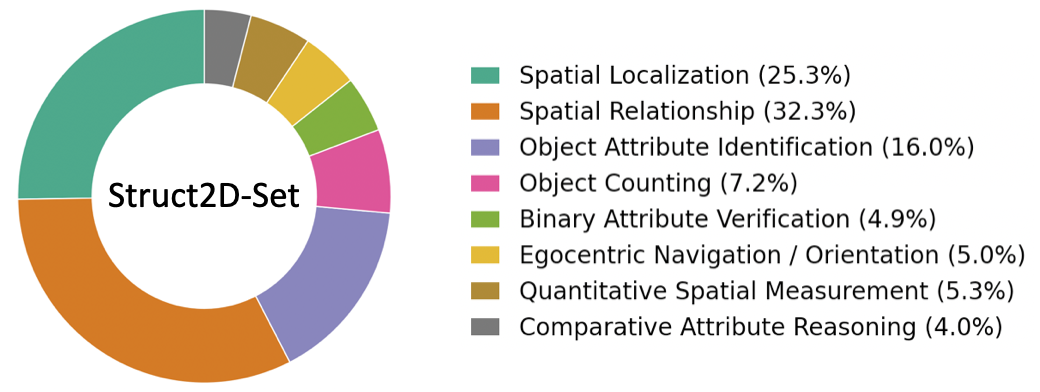}
    \caption{\label{fig:data_dist}\textbf{Distribution of QA types in \data{}.} The dataset covers a diverse range of spatial reasoning skills, with a focus on spatial relationships and localization tasks that require strong geometric understanding.}
    \vspace{-5pt}
\end{wrapfigure}

\textbf{Overview.}
\data{} consists of 200K QA pairs generated from over 6K richly annotated indoor scenes, sourced from large-scale 3D reconstruction datasets—ARKitScenes\cite{baruch2021arkitscenes}, ScanNet~\cite{dai2017scannet}, and ScanNet++\cite{yeshwanth2023scannet++}. These datasets capture diverse real-world environments, including homes, offices, and industrial settings. The QA pairs cover eight categories of spatial reasoning tasks. Figure~\ref{fig:data_dist} shows the distribution of question types across the dataset.

\begin{figure}
    \centering
    \includegraphics[width=\linewidth]{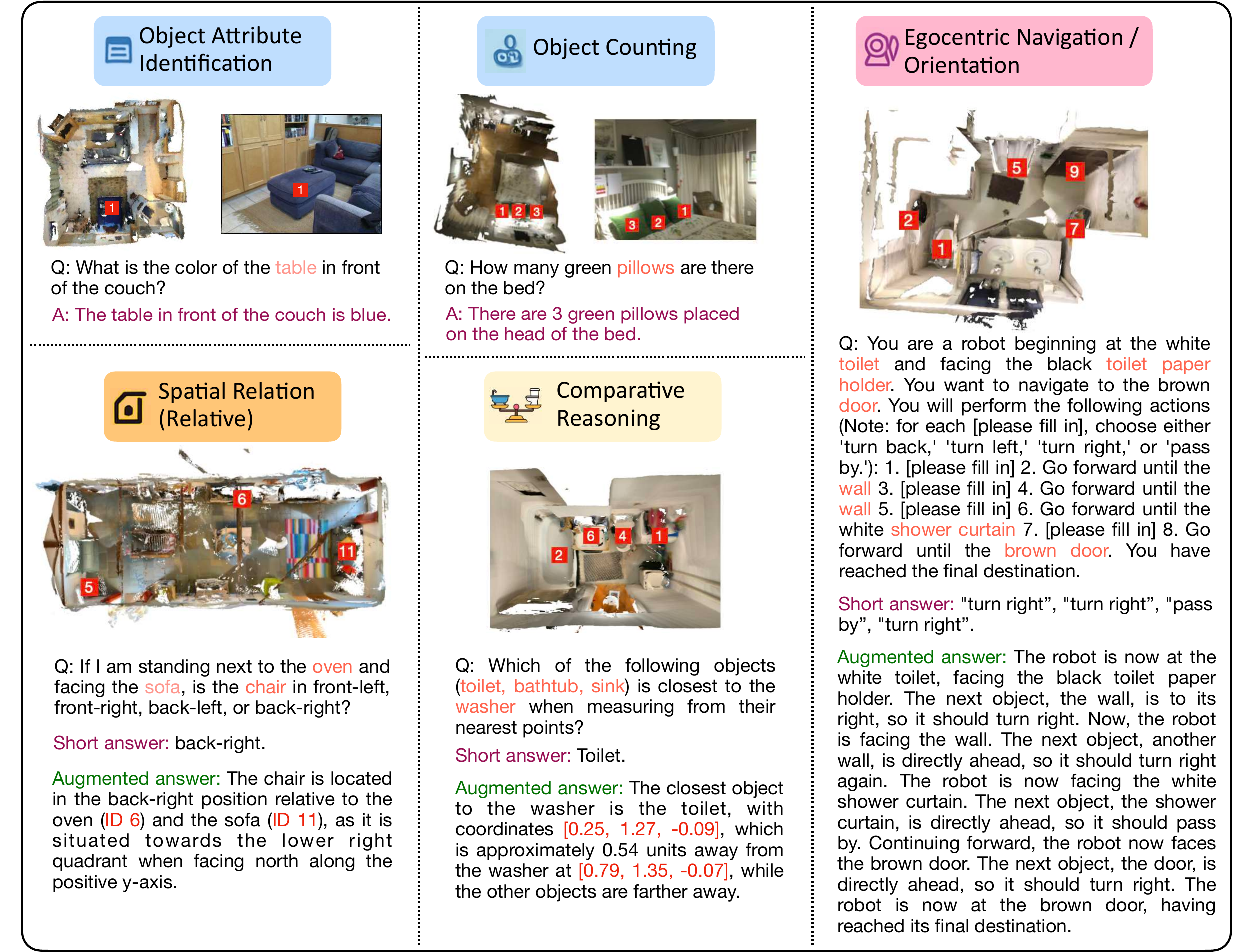}
    \caption{\label{fig:data_example} \textbf{QA examples of \data{}.} Examples cover diverse spatial reasoning tasks, including object attributes, counting, relative positioning, navigation, and comparative reasoning. Each QA pair includes a short answer from 3D geometry and an augmented answer with detailed reasoning generated by ChatGPT.}
\end{figure}

\textbf{Construction pipeline.}
We generate two types of QA pairs to support both spatial reasoning and scene understanding tasks. Each type encompasses multiple subtypes targeting distinct reasoning skills. Representative examples from both types are shown in Figure~\ref{fig:data_example}.

The first type, inspired by VSI-Bench~\cite{yang2024thinking}, involves tasks that require understanding global spatial relationships in 3D, such as spatial relation identification, egocentric navigation, and comparative reasoning. These questions cannot be answered from a single keyframe alone. We begin by extracting ground-truth object annotations from the training sets of the 3D datasets, including object boxes, depth maps, and camera poses. Using structured templates, we generate initial QA pairs based on this meta information, and then enrich them using ChatGPT to produce step-by-step reasoning traces and more natural language formulations. Each QA pair includes a short answer derived from geometry templates and a long-form answer elaborating on the reasoning process. 

The second type of QA pairs is adapted from existing 3D scene understanding benchmarks, including ScanQA~\cite{azuma2022scanqa}, SQA3D~\cite{ma2022sqa3d}, Scan2Cap~\cite{chen2021scan2cap}, ScanRefer~\cite{chen2020scanrefer}, and Multi3DRefer~\cite{Multi3DRefer}. These examples cover tasks such as object attribute identification, counting, and binary verification. We augment the original training set questions and descriptions using ChatGPT to improve clarity and reasoning depth. These tasks typically benefit from selecting keyframes where relevant objects are clearly visible, allowing the model to ground spatial reasoning in egocentric frames.

\subsection{Experiment Setup}
We fine-tune the open-source MLLM Qwen2.5VL~\cite{wang2024qwen2} using our proposed dataset, \data{}. For evaluation, we primarily focus on VSI-Bench~\cite{yang2024thinking}, which includes complex spatial reasoning tasks such as relative direction and route planning. Additionally, we assess model performance on three standard 3D scene understanding tasks built on ScanNet~\cite{dai2017scannet}: 3D question answering (ScanQA~\cite{azuma2022scanqa}, SQA3D~\cite{ma2022sqa3d}), 3D dense captioning (Scan2Cap~\cite{chen2021scan2cap}), and 3D visual grounding (ScanRefer~\cite{chen2020scanrefer}, Multi3DRef~\cite{Multi3DRefer}).
For VSI-Bench, we input only BEV images with filtered object marks and metadata, as the tasks focus purely on spatial relationships. For the other benchmarks, which often involve object attributes or visual details, we additionally provide selected egocentric keyframes to support reasoning.

\subsection{Implementation Details}
We adopt Qwen2.5VL~\cite{wang2024qwen2} as our base MLLM for instruction tuning. During training, the model receives BEV images with filtered object marks and object-centric metadata. For tasks that require appearance or attribute information (e.g., object color or count), we additionally provide egocentric keyframes. All visual inputs are resized to $480 \times 480$, and object marks are adaptively scaled based on their original resolution.

For questions involving complex spatial reasoning, such as relative direction or route planning, we insert special tokens \texttt{<think>} and \texttt{</think>} to guide the model to generate a step-by-step reasoning process, followed by the final answer enclosed within \texttt{<answer>} and \texttt{</answer>}. For simpler questions involving object appearance or quantitative estimation, the model is trained to directly produce short answers without reasoning traces.
We train the model for one epoch using a base learning rate of 2e-6 with cosine annealing. Training with the whole \data{} takes approximately 8 hours on 8×H200 GPUs. 
For evaluation, we follow~\cite{huang2023chat,qi2025gpt4scene} by reconstructing point clouds offline using BundleFusion~\cite{dai2017bundlefusion}, detecting 3D object boxes with Mask3D and UniDet, and projecting them into BEV images and 2D object marks.

\begin{table*}[t]
    \centering
    \caption{ \label{tab:vsi-bench}\textbf{Performance comparison of various models on VSI-Bench~\cite{yang2024thinking}.} The model fine-tuned with \data{} surpasses both the \model{} prompting and the video-based tuning baseline.}
    \vspace{-0.8em}
    \setlength{\tabcolsep}{4pt}
    \resizebox{\linewidth}{!}{
    \begin{tabular}{l c c c c c c c c}
        \toprule
        \multirow{2}{*}{\textbf{Methods}}  & \multirow{2}{*}{\textbf{Avg.}} & \multicolumn{4}{c}{\textbf{Numerical Answer}} & \multicolumn{3}{c}{\textbf{Multiple-Choice Answer}} \\
        \cmidrule(lr){3-6} \cmidrule(lr){7-9}
        & & Obj. Count & Abs. Dist. & Room Size & Obj. Size & Rel. Dist. & Rel. Dir. & Route Plan  \\
        \midrule
        \rowA \multicolumn{9}{l}{\textit{Open-source Models}} \\
        InternVL2-2B~\cite{chen2024internvl}  & 30.3 & 21.8 & 24.9 & 35.0 & 22.0 & 33.8 & 44.2 & 30.5 \\
        InternVL2-8B~\cite{chen2024internvl}  & 33.9 & 23.1 & 28.7&  39.8 & 48.2 & 36.7 & 30.7 & 29.9 \\
        LongVILA-8B~\cite{xue2024longvila}  & 21.1 & 29.1 & 9.1 &  0.0 & 16.7 & 29.6 & 30.7 & 32.5 \\
        VILA-1.5-8B~\cite{lin2023vila}  & 29.5 & 17.4 & 21.8 & 18.8 & 50.3 & 32.1 & 34.8 & 31.0 \\
        LongVA-7B~\cite{zhang2024longva}  & 31.1 & 38.0 & 16.6 & 22.2 & 38.9 & 33.1 & 43.3 & 25.4 \\
        LLaVA-NeXT-Video-7B~\cite{zhang2024llavanext-video}  & 36.3 & \textbf{48.5} & 14.0 & 24.2 & 47.8 & \textbf{43.5} & 42.4 & 34.0 \\
        LLaVA-OneVision-0.5B~\cite{li2024llava}  & 31.2 & 46.1 & 28.4 & 28.3 & 15.4 & 28.9 & 36.9 & 34.5 \\
        LLaVA-OneVision-7B~\cite{li2024llava}  &  33.5 & 47.7 & 20.2 & 12.3 & 47.4 & 42.5 & 35.2 & 29.4 \\
        R1-Zero-VSI~\cite{liao2025improved} (Qwen2-VL-7B) & 32.1 & 39.4 & 25.0 & 43.2 & 25.8 & 32.6 & 30.9 & 27.8 \\
        R1-Zero-VSI~\cite{liao2025improved} (Qwen2-VL-7B) + SFT & 38.8 & 44.7 & 27.6 & \textbf{50.4} & 46.1 & 34.0 & 35.7 & 33.0 \\ 
       \midrule
         \rowA \multicolumn{9}{l}{\textit{Ours}} \\
         Qwen2.5-VL-3B & 25.6 & 27.0 & 22.0 & 25.6 & 32.5 & 17.5 & 28.9 & 25.6 \\
         Qwen2.5-VL-3B~(\model{} Prompting ) & 29.4 & 46.6 & 24.6 & 22.3 & 33.6 & 21.2 & 30.5 & 27.2 \\
         Qwen2.5-VL-3B~(Baseline) & 33.9 & 24.6 & 34.0 & 46.4 & 53.5 & 21.2 & 30.5 & 27.2 \\
         Qwen2.5-VL-3B~(SFT) & 41.9 & 46.0 & 34.7 & 42.6 & 56.4 & 35.1 & 44.9 & 33.5 \\
         Qwen2.5-VL-7B~(SFT) & \textbf{43.6} & 47.1 & \textbf{35.1} & 48.9 & \textbf{57.1} & 35.1 & \textbf{45.9} & \textbf{35.8} \\
        \bottomrule
    \end{tabular}
    }
    \vspace{-5pt}
    
\end{table*}

\begin{table*}[t]
\centering
\vspace{-2mm}
\caption{\label{tab:scanqa}\textbf{3D Question Answering Evaluation on ScanQA~\cite{azuma2022scanqa} and SQA3D~\cite{ma2022sqa3d} datasets.}}
\vspace{-0.8em}
\resizebox{\linewidth}{!}{
\begin{tabular}{lccccccc}
\toprule
\multirow{2}{*}{\textbf{Methods}} & \multicolumn{5}{c}{\textbf{ScanQA(val)}} & \multicolumn{2}{c}{\textbf{SQA3D(val)}} \\
        \cmidrule(lr){2-6} \cmidrule(lr){7-8}
        &  BLEU-1 & BLEU-4 & METEOR & ROUGE & CIDEr & EM-1 & EM-R1 \\
\midrule
\rowA \multicolumn{8}{l}{\textit{Task-Specific Model}} \\
ScanQA~\cite{azuma2022scanqa}  & 30.2 & 10.1 & 13.1 & 33.3 & 64.9 & -- & -- \\
SQ3D~\cite{ma2022sqa3d}  & -- & -- & -- & -- & -- & 46.6 & -- \\
3D-VLP~\cite{jin2023context}  & 30.5 & 11.2 & 13.5 & 34.5 & -- & -- & -- \\
3D-Vista~\cite{zhu20233d}  & -- & -- & 13.9 & 35.7 & -- & 48.5 & -- \\
\midrule
\rowA \multicolumn{8}{l}{\textit{3D LLM Based Model}} \\
Chat-3D~\cite{wang2023chat}  & 29.1 & 6.4 & 11.9 & 28.5 & 53.2 & -- & -- \\
Chat-3D v2~\cite{huang2023chat} & 38.4 & 7.3 & 16.1 & 40.1 & 77.1 & -- & -- \\
3D-LLM~\cite{hong20233d}  & 39.3 & 12.0 & 14.5 & 37.3 & 69.4 & -- & -- \\
LL3DA~\cite{chen2024ll3da} & -- & 13.5 & 15.9 & 37.3 & 76.8 & -- & -- \\
PQ3D~\cite{zhu2024unifying}  & -- & -- & -- & -- & -- & 47.1 & -- \\
LEO~\cite{huang2023embodied}  & -- & 11.5 & 16.2 & 39.3 & 80.0 & 50.0 & 50.0 \\
Chat-Scene~\cite{huang2023chat}  & 43.2 & 14.3 & 18.0 & 41.6 & 87.7 & 54.6 & 57.5 \\
\midrule
\rowA \multicolumn{8}{l}{\textit{Vision LLM Based Model}} \\
InternVL-2-8B~\cite{chen2024internvl}  & 23.9 & 3.3 & 14.5 & 34.3 & 62.5 & 33.0 & 45.3 \\
MiniCPM-V-2.6~\cite{yao2024minicpm}  & 25.1 & 8.4 & 11.8 & 31.5 & 60.1 & 42.6 & 46.6 \\
Qwen2-VL-7B (GPT4Scene)  & 43.4 & 14.6 & \textbf{17.7} & 43.6 & 90.9 & 57.4 & 60.7 \\
Qwen2.5-VL-7B (Ours)  & \textbf{45.2} & \textbf{15.8} & 17.4 & \textbf{44.1} & \textbf{92.1} & \textbf{58.5} & \textbf{61.3} \\
\bottomrule
\end{tabular}
}
\vspace{-15pt}
\end{table*}

\begin{table}
    \centering
    \caption{\label{tab:ablation} \textbf{Ablation on different variants.} To save computational resource, models are trained with Qwen2.5VL-3B model by default.}
    \resizebox{\linewidth}{!}{
    \begin{tabular}{l| c c c c c c c c c}
        \toprule
        \multirow{2}{*}{\textbf{Settings}}  & \multirow{2}{*}{\textbf{Avg.}} & \multicolumn{3}{c}{\textbf{Numerical Answer}} & \multicolumn{3}{c}{\textbf{Multiple-Choice Answer}} \\
        \cmidrule(lr){3-5} \cmidrule(lr){6-8}
        & &  Obj. Count & Abs. Dist. & Room Size & Rel. Dist. & Rel. Dir. & Route Plan \\
        \midrule
        \rowA \multicolumn{8}{l}{\textit{Tuning Data Format}} \\
           wo/ augmented QA  & 31.5 & 43.7 & 33.1 & 34.1 & 32.1 & 14.7 & 31.5  \\
     w/ augmented QA  &   38.0 & 44.4 & 33.6 & 41.5 & 33.3 & 42.2 & 33.0 \\
     \midrule
     \rowA \multicolumn{8}{l}{\textit{Evaluation Strategy}} \\
    wo/ \texttt{</think>}  & 36.2 & 44.1 & 33.6 & 41.5 & 33.3 & 38.6 & 26.3 \\
    w/ \texttt{</think>} &   36.1 & 44.4 & 30.0 & 35.6 & 31.5 & 42.2 & 33.0  \\
        \bottomrule
    \end{tabular}
    }  
    \vspace{-20pt}
\end{table}

\subsection{Main results}
We present quantitative results on VSI-Bench\cite{yang2024thinking} in Table~\ref{tab:vsi-bench} and on ScanQA\cite{azuma2022scanqa} and SQA3~\cite{ma2022sqa3d} in Table~\ref{tab:scanqa}. Additional benchmark results are provided in the Appendix due to space limitations.

As shown in Table~\ref{tab:vsi-bench}, our model fine-tuned with the \data{} dataset achieves the highest average score (43.6) among all open-source models evaluated on VSI-Bench. Notably, it surpasses both the \model{} prompting variant (29.4) and the standard baseline trained with uniformly sampled 16 video frames (33.9), confirming the effectiveness of our full instruction tuning approach. The performance gains are especially prominent on spatial reasoning tasks like relative direction (45.9) and route planning (35.8), where the model must integrate both geometric understanding and egocentric context. Compared with R1-Zero-VSI~\cite{liao2025improved} (38.8), a recent method that trains Qwen2-VL-7B using video-based supervision, our tuned model not only achieves stronger average performance but also uses fewer visual frames and does not rely on dense temporal input. These results highlight the scalability and efficiency of \data{} for training capable spatial reasoners without explicit 3D features at inference.

Table~\ref{tab:scanqa} shows results on two traditional 3D question answering benchmarks, ScanQA and SQA3D. Our model outperforms most existing methods, including several that rely on explicit 3D point cloud inputs. Compared with GPT4Scene~\cite{qi2025gpt4scene}, our model performs on par across most metrics. However, these benchmarks primarily require identifying relevant keyframes and generating free-form textual answers. As a result, models can often rely on memorizing object-level attributes, and the rule-based evaluation metrics (\eg, BLEU, CIDEr) may not fully reflect the correctness or reasoning quality of the generated answers.
Please refer to Appendix for qualitative results.
\begin{figure}
    \centering
    \includegraphics[width=\linewidth]{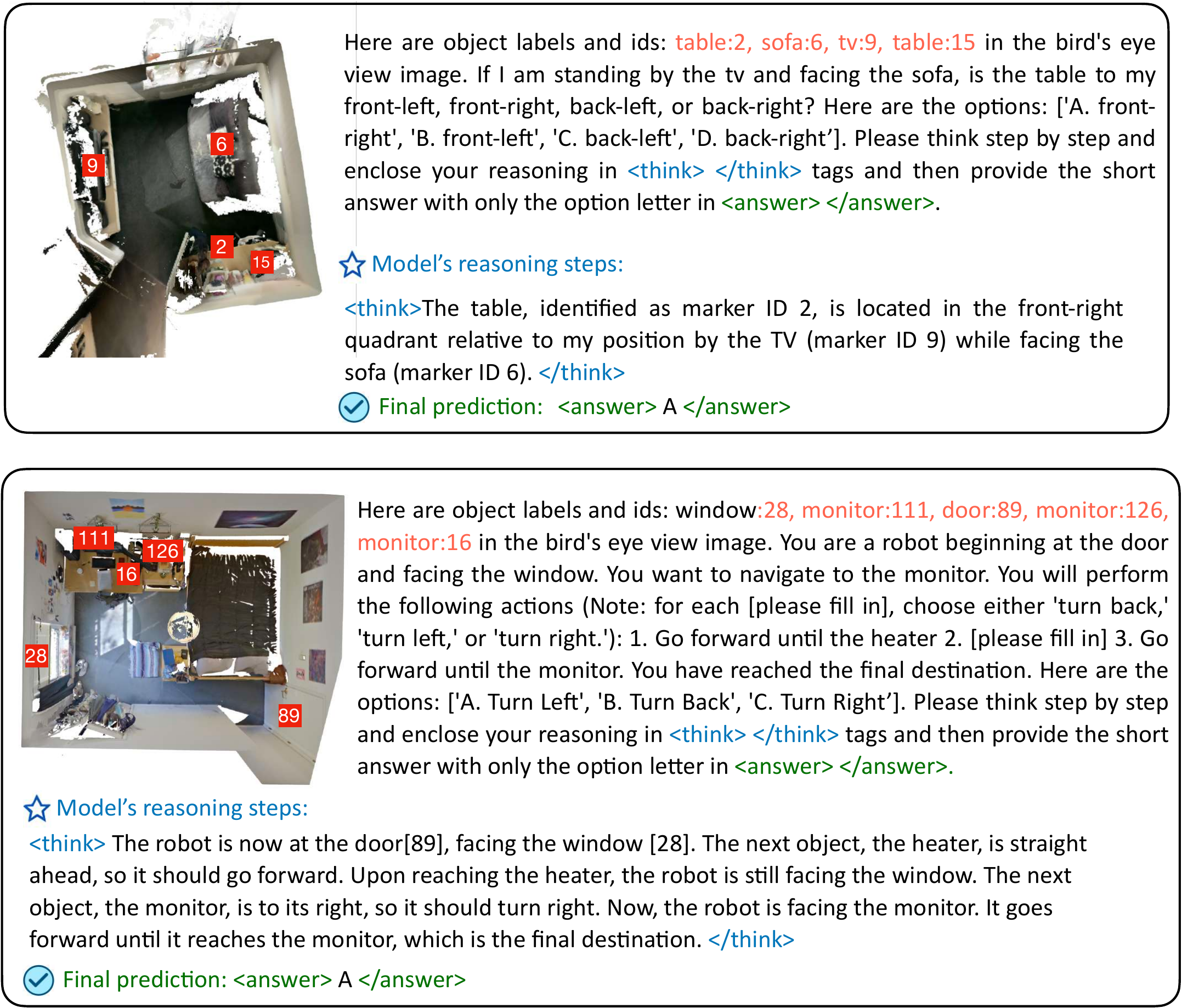}
    \caption{\label{fig:data_example_think}\textbf{Example reasoning traces using \texttt{<think>} and \texttt{<answer>} prompting.}
The top example shows relative direction reasoning, where the model infers the spatial relation between objects from a fixed viewpoint in the BEV image. The bottom example shows step-by-step planning, requiring the model to simulate orientation changes across a sequence of waypoints.}
    \vspace{-15pt}
\end{figure}
\textbf{Ablation Study}
To better understand the impact of individual components in our framework, we conduct a series of ablation studies using the Qwen2.5VL-3B model for efficiency, as shown in Table~\ref{tab:ablation}.
First, we evaluate the effect of QA augmentation. Incorporating enriched QA pairs generated with ChatGPT leads to a substantial improvement in overall performance (Avg: 38.0 vs. 31.5), especially on reasoning-heavy tasks such as relative direction (42.2 vs. 14.7). This supports our earlier claim that step-by-step reasoning traces help guide the model's attention and inference.
We further assess the role of explicit reasoning supervision using the \texttt{<think>} and \texttt{<answer>} tokens. While the average scores are comparable, including \texttt{<think>} tokens improves performance on reasoning-intensive tasks like relative direction and route planning (42.2 vs. 38.6 and 33.0 vs. 26.3, respectively), indicating that instructing the model to explicitly reason can enhance output quality on complex spatial questions.
Figure~\ref{fig:data_example_think} illustrates step-by-step reasoning generated using our \texttt{<think>} and \texttt{<answer>} prompting format in egocentric navigation tasks. In each case, the model is prompted to analyze object positions and spatial transitions from a top-down BEV image with object markers. These examples highlight the model's ability to decompose spatial reasoning tasks into interpretable steps and generate structured answers grounded in visual context.

\section{Limitations and future work}
While \model{} demonstrates strong spatial reasoning capabilities with structured 2D inputs, there remain areas where future work could extend its applicability:

\begin{itemize}[nosep,leftmargin=2em]
\item \textbf{3D preprocessing requirements.} Although \model{} does not use 3D features during inference, it currently relies on 3D perception modules to generate BEV images and object-centric metadata. This may pose a challenge in latency-sensitive or resource-constrained environments. However, since \model{} is agnostic to the specific perception backbone, it can readily integrate with ongoing advances in real-time and lightweight 3D reconstruction systems.
\item \textbf{Indoor scene focus.} The current version of \data{} is constructed from over 6K richly annotated indoor scenes, including homes, offices, and classrooms. While this enables detailed reasoning in structured environments, generalization to outdoor or open-world scenes remains less explored. Incorporating diverse spatial layouts and object categories from outdoor domains is a promising direction for future dataset expansion.

\end{itemize}

\section{Conclusion}

We present \model{}, a perception-guided framework that enables MLLMs to perform 3D spatial reasoning using structured 2D inputs. Through zero-shot analysis and instruction tuning, we show that BEV images, object-centric metadata, and keyframes are sufficient to unlock strong spatial reasoning capabilities—without requiring explicit 3D inputs.
Our curated dataset, \data{}, supports scalable instruction tuning with fine-grained QA pairs grounded in real 3D scenes. Fine-tuning with \data{} yields significant gains across spatial reasoning benchmarks, outperforming prior open-source methods even under noisy perception.
These findings demonstrate that structured 2D projections are a practical and effective alternative to direct 3D representations, offering a scalable path toward robust multimodal spatial understanding in MLLMs.

\section*{Acknowledgment}

We thank Zhangyang Qi for the thoughtful discussions.

\bibliographystyle{natbib}
\bibliography{main}

@String(CVPR= {IEEE Conf. Comput. Vis. Pattern Recog.})

@String(ICCV= {Int. Conf. Comput. Vis.})

@String(ECCV= {Eur. Conf. Comput. Vis.})

@String(NIPS= {Adv. Neural Inform. Process. Syst.})

@String(TOG= {ACM Trans. Graph.})

@String(ICLR = {Int. Conf. Learn. Represent.})

@String(CVPR  = {CVPR})

@String(ICCV  = {ICCV})

@String(ECCV  = {ECCV})

@String(NIPS  = {NeurIPS})

@String(TOG   = {ACM TOG})

@String(ICLR  = {ICLR})

@article{hong20233d,
  title={3d-llm: Injecting the 3d world into large language models},
  author={Hong, Yining and Zhen, Haoyu and Chen, Peihao and Zheng, Shuhong and Du, Yilun and Chen, Zhenfang and Gan, Chuang},
  journal=NIPS,
  year={2023}
}

@article{qi2025gpt4scene,
  title={GPT4Scene: Understand 3D Scenes from Videos with Vision-Language Models},
  author={Qi, Zhangyang and Zhang, Zhixiong and Fang, Ye and Wang, Jiaqi and Zhao, Hengshuang},
  journal={arXiv preprint arXiv:2501.01428},
  year={2025}
}

@article{yang2024thinking,
  title={Thinking in space: How multimodal large language models see, remember, and recall spaces},
  author={Yang, Jihan and Yang, Shusheng and Gupta, Anjali W and Han, Rilyn and Fei-Fei, Li and Xie, Saining},
  journal={arXiv preprint arXiv:2412.14171},
  year={2024}
}

@article{huang2023chat,
  title={Chat-Scene: Bridging 3D Scene and Large Language Models with Object Identifiers},
  author={Huang, Haifeng and Chen, Yilun and Wang, Zehan and Huang, Rongjie and Xu, Runsen and Wang, Tai and Liu, Luping and Cheng, Xize and Zhao, Yang and Pang, Jiangmiao and others},
  journal={arXiv preprint arXiv:2312.08168},
  year={2023}
}

@article{yang2023set,
  title={Set-of-mark prompting unleashes extraordinary visual grounding in gpt-4v},
  author={Yang, Jianwei and Zhang, Hao and Li, Feng and Zou, Xueyan and Li, Chunyuan and Gao, Jianfeng},
  journal={arXiv preprint arXiv:2310.11441},
  year={2023}
}

@article{yang2023dawn,
  title={The dawn of lmms: Preliminary explorations with gpt-4v (ision)},
  author={Yang, Zhengyuan and Li, Linjie and Lin, Kevin and Wang, Jianfeng and Lin, Chung-Ching and Liu, Zicheng and Wang, Lijuan},
  journal={arXiv preprint arXiv:2309.17421},
  volume={9},
  number={1},
  pages={1},
  year={2023}
}

@article{cai2024spatialbot,
  title={Spatialbot: Precise spatial understanding with vision language models},
  author={Cai, Wenxiao and Ponomarenko, Iaroslav and Yuan, Jianhao and Li, Xiaoqi and Yang, Wankou and Dong, Hao and Zhao, Bo},
  journal={arXiv preprint arXiv:2406.13642},
  year={2024}
}

@inproceedings{chen2024spatialvlm,
  title={Spatialvlm: Endowing vision-language models with spatial reasoning capabilities},
  author={Chen, Boyuan and Xu, Zhuo and Kirmani, Sean and Ichter, Brain and Sadigh, Dorsa and Guibas, Leonidas and Xia, Fei},
  booktitle={Proceedings of the IEEE/CVF Conference on Computer Vision and Pattern Recognition},
  pages={14455--14465},
  year={2024}
}

@article{cheng2024spatialrgpt,
  title={SpatialRGPT: Grounded Spatial Reasoning in Vision Language Models},
  author={Cheng, An-Chieh and Yin, Hongxu and Fu, Yang and Guo, Qiushan and Yang, Ruihan and Kautz, Jan and Wang, Xiaolong and Liu, Sifei},
  journal={arXiv preprint arXiv:2406.01584},
  year={2024}
}

@article{li2024topviewrs,
  title={Topviewrs: Vision-language models as top-view spatial reasoners},
  author={Li, Chengzu and Zhang, Caiqi and Zhou, Han and Collier, Nigel and Korhonen, Anna and Vuli{\'c}, Ivan},
  journal={arXiv preprint arXiv:2406.02537},
  year={2024}
}

@article{liu2024coarse,
  title={Coarse correspondence elicit 3d spacetime understanding in multimodal language model},
  author={Liu, Benlin and Dong, Yuhao and Wang, Yiqin and Rao, Yongming and Tang, Yansong and Ma, Wei-Chiu and Krishna, Ranjay},
  journal={arXiv preprint arXiv:2408.00754},
  year={2024}
}

@inproceedings{yang2024v,
  title={V-irl: Grounding virtual intelligence in real life},
  author={Yang, Jihan and Ding, Runyu and Brown, Ellis and Qi, Xiaojuan and Xie, Saining},
  booktitle={European Conference on Computer Vision},
  pages={36--55},
  year={2024},
  organization={Springer}
}

@article{zhu2024llava,
  title={Llava-3d: A simple yet effective pathway to empowering lmms with 3d-awareness},
  author={Zhu, Chenming and Wang, Tai and Zhang, Wenwei and Pang, Jiangmiao and Liu, Xihui},
  journal={arXiv preprint arXiv:2409.18125},
  year={2024}
}

@article{ramakrishnan2024does,
  title={Does Spatial Cognition Emerge in Frontier Models?},
  author={Ramakrishnan, Santhosh Kumar and Wijmans, Erik and Kraehenbuehl, Philipp and Koltun, Vladlen},
  journal={arXiv preprint arXiv:2410.06468},
  year={2024}
}

@article{tang2024sparkle,
  title={Sparkle: Mastering basic spatial capabilities in vision language models elicits generalization to composite spatial reasoning},
  author={Tang, Yihong and Qu, Ao and Wang, Zhaokai and Zhuang, Dingyi and Wu, Zhaofeng and Ma, Wei and Wang, Shenhao and Zheng, Yunhan and Zhao, Zhan and Zhao, Jinhua},
  journal={arXiv preprint arXiv:2410.16162},
  year={2024}
}

@article{momennejad2023evaluating,
  title={Evaluating cognitive maps and planning in large language models with cogeval},
  author={Momennejad, Ida and Hasanbeig, Hosein and Vieira Frujeri, Felipe and Sharma, Hiteshi and Jojic, Nebojsa and Palangi, Hamid and Ness, Robert and Larson, Jonathan},
  journal={Advances in Neural Information Processing Systems},
  volume={36},
  pages={69736--69751},
  year={2023}
}

@article{rozanova2021grounding,
  title={Grounding Natural Language Instructions: Can Large Language Models Capture Spatial Information?},
  author={Rozanova, Julia and Ferreira, Deborah and Dubba, Krishna and Cheng, Weiwei and Zhang, Dell and Freitas, Andre},
  journal={arXiv preprint arXiv:2109.08634},
  year={2021}
}

@article{wu2024visualization,
  title={Visualization-of-thought elicits spatial reasoning in large language models},
  author={Wu, Wenshan and Mao, Shaoguang and Zhang, Yadong and Xia, Yan and Dong, Li and Cui, Lei and Wei, Furu},
  journal={arXiv e-prints},
  pages={arXiv--2404},
  year={2024}
}

@article{marsili2025visual,
  title={Visual Agentic AI for Spatial Reasoning with a Dynamic API},
  author={Marsili, Damiano and Agrawal, Rohun and Yue, Yisong and Gkioxari, Georgia},
  journal={arXiv preprint arXiv:2502.06787},
  year={2025}
}

@article{guo2023point,
  title={Point-bind \& point-llm: Aligning point cloud with multi-modality for 3d understanding, generation, and instruction following},
  author={Guo, Ziyu and Zhang, Renrui and Zhu, Xiangyang and Tang, Yiwen and Ma, Xianzheng and Han, Jiaming and Chen, Kexin and Gao, Peng and Li, Xianzhi and Li, Hongsheng and others},
  journal={arXiv preprint arXiv:2309.00615},
  year={2023}
}

@inproceedings{qi2024shapellm,
  title={Shapellm: Universal 3d object understanding for embodied interaction},
  author={Qi, Zekun and Dong, Runpei and Zhang, Shaochen and Geng, Haoran and Han, Chunrui and Ge, Zheng and Yi, Li and Ma, Kaisheng},
  booktitle={European Conference on Computer Vision},
  pages={214--238},
  year={2024},
  organization={Springer}
}

@inproceedings{qi2024gpt4point,
  title={Gpt4point: A unified framework for point-language understanding and generation},
  author={Qi, Zhangyang and Fang, Ye and Sun, Zeyi and Wu, Xiaoyang and Wu, Tong and Wang, Jiaqi and Lin, Dahua and Zhao, Hengshuang},
  booktitle={Proceedings of the IEEE/CVF Conference on Computer Vision and Pattern Recognition},
  pages={26417--26427},
  year={2024}
}

@inproceedings{xu2024pointllm,
  title={Pointllm: Empowering large language models to understand point clouds},
  author={Xu, Runsen and Wang, Xiaolong and Wang, Tai and Chen, Yilun and Pang, Jiangmiao and Lin, Dahua},
  booktitle={European Conference on Computer Vision},
  pages={131--147},
  year={2024},
  organization={Springer}
}

@inproceedings{X-trans2cap,
  title={X-trans2cap: Cross-modal knowledge transfer using transformer for 3d dense captioning},
  author={Yuan, Zhihao and Yan, Xu and Liao, Yinghong and Guo, Yao and Li, Guanbin and Cui, Shuguang and Li, Zhen},
  booktitle=CVPR,
  year={2022}
}

@inproceedings{3djcg,
  title={3djcg: A unified framework for joint dense captioning and visual grounding on 3d point clouds},
  author={Cai, Daigang and Zhao, Lichen and Zhang, Jing and Sheng, Lu and Xu, Dong},
  booktitle={CVPR},
  year={2022}
}

@inproceedings{vote2cap-detr,
  title={End-to-end 3d dense captioning with vote2cap-detr},
  author={Chen, Sijin and Zhu, Hongyuan and Chen, Xin and Lei, Yinjie and Yu, Gang and Chen, Tao},
  booktitle=CVPR,
  year={2023}
}

@article{fu2024scene,
  title={Scene-llm: Extending language model for 3d visual understanding and reasoning},
  author={Fu, Rao and Liu, Jingyu and Chen, Xilun and Nie, Yixin and Xiong, Wenhan},
  journal={arXiv preprint arXiv:2403.11401},
  year={2024}
}

@article{man2024lexicon3d,
  title={Lexicon3d: Probing visual foundation models for complex 3d scene understanding},
  author={Man, Yunze and Zheng, Shuhong and Bao, Zhipeng and Hebert, Martial and Gui, Liang-Yan and Wang, Yu-Xiong},
  journal={arXiv preprint arXiv:2409.03757},
  year={2024}
}

@article{sun2024layoutvlm,
  title={LayoutVLM: Differentiable Optimization of 3D Layout via Vision-Language Models},
  author={Sun, Fan-Yun and Liu, Weiyu and Gu, Siyi and Lim, Dylan and Bhat, Goutam and Tombari, Federico and Li, Manling and Haber, Nick and Wu, Jiajun},
  journal={arXiv preprint arXiv:2412.02193},
  year={2024}
}

@article{liu20233daxiesprompts,
  title={3daxiesprompts: Unleashing the 3d spatial task capabilities of gpt-4v},
  author={Liu, Dingning and Dong, Xiaomeng and Zhang, Renrui and Luo, Xu and Gao, Peng and Huang, Xiaoshui and Gong, Yongshun and Wang, Zhihui},
  journal={arXiv preprint arXiv:2312.09738},
  year={2023}
}

@inproceedings{chen2024ll3da,
  title={Ll3da: Visual interactive instruction tuning for omni-3d understanding reasoning and planning},
  author={Chen, Sijin and Chen, Xin and Zhang, Chi and Li, Mingsheng and Yu, Gang and Fei, Hao and Zhu, Hongyuan and Fan, Jiayuan and Chen, Tao},
  booktitle={Proceedings of the IEEE/CVF Conference on Computer Vision and Pattern Recognition},
  pages={26428--26438},
  year={2024}
}

@article{zheng2024tracevla,
  title={Tracevla: Visual trace prompting enhances spatial-temporal awareness for generalist robotic policies},
  author={Zheng, Ruijie and Liang, Yongyuan and Huang, Shuaiyi and Gao, Jianfeng and Daum{\'e} III, Hal and Kolobov, Andrey and Huang, Furong and Yang, Jianwei},
  journal={arXiv preprint arXiv:2412.10345},
  year={2024}
}

@article{zhang2024earthmarker,
  title={Earthmarker: A visual prompting multi-modal large language model for remote sensing},
  author={Zhang, Wei and Cai, Miaoxin and Zhang, Tong and Zhuang, Yin and Li, Jun and Mao, Xuerui},
  journal={IEEE Transactions on Geoscience and Remote Sensing},
  year={2024},
  publisher={IEEE}
}

@article{lee2024affordance,
  title={Affordance-guided reinforcement learning via visual prompting},
  author={Lee, Olivia Y and Xie, Annie and Fang, Kuan and Pertsch, Karl and Finn, Chelsea},
  journal={arXiv preprint arXiv:2407.10341},
  year={2024}
}

@article{wu2024visual,
  title={Visual prompting in multimodal large language models: A survey},
  author={Wu, Junda and Zhang, Zhehao and Xia, Yu and Li, Xintong and Xia, Zhaoyang and Chang, Aaron and Yu, Tong and Kim, Sungchul and Rossi, Ryan A and Zhang, Ruiyi and others},
  journal={arXiv preprint arXiv:2409.15310},
  year={2024}
}

@article{lin2024draw,
  title={Draw-and-understand: Leveraging visual prompts to enable mllms to comprehend what you want},
  author={Lin, Weifeng and Wei, Xinyu and An, Ruichuan and Gao, Peng and Zou, Bocheng and Luo, Yulin and Huang, Siyuan and Zhang, Shanghang and Li, Hongsheng},
  journal={arXiv preprint arXiv:2403.20271},
  year={2024}
}

@article{hu2025leveraging,
  title={Leveraging hallucinations to reduce manual prompt dependency in promptable segmentation},
  author={Hu, Jian and Lin, Jiayi and Yan, Junchi and Gong, Shaogang},
  journal={Advances in Neural Information Processing Systems},
  volume={37},
  pages={107171--107197},
  year={2025}
}

@inproceedings{yu2024attention,
  title={Attention prompting on image for large vision-language models},
  author={Yu, Runpeng and Yu, Weihao and Wang, Xinchao},
  booktitle={European Conference on Computer Vision},
  pages={251--268},
  year={2024},
  organization={Springer}
}

@inproceedings{cai2024vip,
  title={Vip-llava: Making large multimodal models understand arbitrary visual prompts},
  author={Cai, Mu and Liu, Haotian and Mustikovela, Siva Karthik and Meyer, Gregory P and Chai, Yuning and Park, Dennis and Lee, Yong Jae},
  booktitle={Proceedings of the IEEE/CVF Conference on Computer Vision and Pattern Recognition},
  pages={12914--12923},
  year={2024}
}

@article{hu2025visual,
  title={Visual sketchpad: Sketching as a visual chain of thought for multimodal language models},
  author={Hu, Yushi and Shi, Weijia and Fu, Xingyu and Roth, Dan and Ostendorf, Mari and Zettlemoyer, Luke and Smith, Noah A and Krishna, Ranjay},
  journal={Advances in Neural Information Processing Systems},
  volume={37},
  pages={139348--139379},
  year={2025}
}

@article{nasiriany2024pivot,
  title={Pivot: Iterative visual prompting elicits actionable knowledge for vlms},
  author={Nasiriany, Soroush and Xia, Fei and Yu, Wenhao and Xiao, Ted and Liang, Jacky and Dasgupta, Ishita and Xie, Annie and Driess, Danny and Wahid, Ayzaan and Xu, Zhuo and others},
  journal={arXiv preprint arXiv:2402.07872},
  year={2024}
}

@article{wu2025controlmllm,
  title={Controlmllm: Training-free visual prompt learning for multimodal large language models},
  author={Wu, Mingrui and Cai, Xinyue and Ji, Jiayi and Li, Jiale and Huang, Oucheng and Luo, Gen and Fei, Hao and Jiang, Guannan and Sun, Xiaoshuai and Ji, Rongrong},
  journal={Advances in Neural Information Processing Systems},
  volume={37},
  pages={45206--45234},
  year={2025}
}

@inproceedings{wu2024dettoolchain,
  title={Dettoolchain: A new prompting paradigm to unleash detection ability of mllm},
  author={Wu, Yixuan and Wang, Yizhou and Tang, Shixiang and Wu, Wenhao and He, Tong and Ouyang, Wanli and Torr, Philip and Wu, Jian},
  booktitle={European Conference on Computer Vision},
  pages={164--182},
  year={2024},
  organization={Springer}
}

@article{dubey2024llama,
  title={The llama 3 herd of models},
  author={Dubey, Abhimanyu and Jauhri, Abhinav and Pandey, Abhinav and Kadian, Abhishek and Al-Dahle, Ahmad and Letman, Aiesha and Mathur, Akhil and Schelten, Alan and Yang, Amy and Fan, Angela and others},
  journal={arXiv preprint arXiv:2407.21783},
  year={2024}
}

@article{chowdhery2023palm,
  title={Palm: Scaling language modeling with pathways},
  author={Chowdhery, Aakanksha and Narang, Sharan and Devlin, Jacob and Bosma, Maarten and Mishra, Gaurav and Roberts, Adam and Barham, Paul and Chung, Hyung Won and Sutton, Charles and Gehrmann, Sebastian and others},
  journal={Journal of Machine Learning Research},
  volume={24},
  number={240},
  pages={1--113},
  year={2023}
}

@article{touvron2023llama,
  title={Llama: Open and efficient foundation language models},
  author={Touvron, Hugo and Lavril, Thibaut and Izacard, Gautier and Martinet, Xavier and Lachaux, Marie-Anne and Lacroix, Timoth{\'e}e and Rozi{\`e}re, Baptiste and Goyal, Naman and Hambro, Eric and Azhar, Faisal and others},
  journal={arXiv preprint arXiv:2302.13971},
  year={2023}
}

@article{zhang2022opt,
  title={Opt: Open pre-trained transformer language models},
  author={Zhang, Susan and Roller, Stephen and Goyal, Naman and Artetxe, Mikel and Chen, Moya and Chen, Shuohui and Dewan, Christopher and Diab, Mona and Li, Xian and Lin, Xi Victoria and others},
  journal={arXiv preprint arXiv:2205.01068},
  year={2022}
}

@article{achiam2023gpt,
  title={Gpt-4 technical report},
  author={Achiam, Josh and Adler, Steven and Agarwal, Sandhini and Ahmad, Lama and Akkaya, Ilge and Aleman, Florencia Leoni and Almeida, Diogo and Altenschmidt, Janko and Altman, Sam and Anadkat, Shyamal and others},
  journal={arXiv preprint arXiv:2303.08774},
  year={2023}
}

@article{hurst2024gpt,
  title={Gpt-4o system card},
  author={Hurst, Aaron and Lerer, Adam and Goucher, Adam P and Perelman, Adam and Ramesh, Aditya and Clark, Aidan and Ostrow, AJ and Welihinda, Akila and Hayes, Alan and Radford, Alec and others},
  journal={arXiv preprint arXiv:2410.21276},
  year={2024}
}

@article{brown2020language,
  title={Language models are few-shot learners},
  author={Brown, Tom and Mann, Benjamin and Ryder, Nick and Subbiah, Melanie and Kaplan, Jared D and Dhariwal, Prafulla and Neelakantan, Arvind and Shyam, Pranav and Sastry, Girish and Askell, Amanda and others},
  journal={Advances in neural information processing systems},
  volume={33},
  pages={1877--1901},
  year={2020}
}

@article{dong2022survey,
  title={A survey on in-context learning},
  author={Dong, Qingxiu and Li, Lei and Dai, Damai and Zheng, Ce and Ma, Jingyuan and Li, Rui and Xia, Heming and Xu, Jingjing and Wu, Zhiyong and Liu, Tianyu and others},
  journal={arXiv preprint arXiv:2301.00234},
  year={2022}
}

@article{wei2022chain,
  title={Chain-of-thought prompting elicits reasoning in large language models},
  author={Wei, Jason and Wang, Xuezhi and Schuurmans, Dale and Bosma, Maarten and Xia, Fei and Chi, Ed and Le, Quoc V and Zhou, Denny and others},
  journal={Advances in neural information processing systems},
  volume={35},
  pages={24824--24837},
  year={2022}
}

@article{yao2023tree,
  title={Tree of thoughts: Deliberate problem solving with large language models},
  author={Yao, Shunyu and Yu, Dian and Zhao, Jeffrey and Shafran, Izhak and Griffiths, Tom and Cao, Yuan and Narasimhan, Karthik},
  journal={Advances in neural information processing systems},
  volume={36},
  pages={11809--11822},
  year={2023}
}

@article{liu2023visual,
  title={Visual instruction tuning},
  author={Liu, Haotian and Li, Chunyuan and Wu, Qingyang and Lee, Yong Jae},
  journal={Advances in neural information processing systems},
  volume={36},
  pages={34892--34916},
  year={2023}
}

@article{zhu2023minigpt,
  title={Minigpt-4: Enhancing vision-language understanding with advanced large language models},
  author={Zhu, Deyao and Chen, Jun and Shen, Xiaoqian and Li, Xiang and Elhoseiny, Mohamed},
  journal={arXiv preprint arXiv:2304.10592},
  year={2023}
}

@article{zhang2023llama,
  title={Llama-adapter: Efficient fine-tuning of language models with zero-init attention},
  author={Zhang, Renrui and Han, Jiaming and Liu, Chris and Gao, Peng and Zhou, Aojun and Hu, Xiangfei and Yan, Shilin and Lu, Pan and Li, Hongsheng and Qiao, Yu},
  journal={arXiv preprint arXiv:2303.16199},
  year={2023}
}

@inproceedings{
dai2023instructblip,
title={Instruct{BLIP}: Towards General-purpose Vision-Language Models with Instruction Tuning},
author={Wenliang Dai and Junnan Li and Dongxu Li and Anthony Tiong and Junqi Zhao and Weisheng Wang and Boyang Li and Pascale Fung and Steven Hoi},
booktitle={Thirty-seventh Conference on Neural Information Processing Systems},
year={2023},
}

@article{li2024multimodal,
  title={Multimodal foundation models: From specialists to general-purpose assistants},
  author={Li, Chunyuan and Gan, Zhe and Yang, Zhengyuan and Yang, Jianwei and Li, Linjie and Wang, Lijuan and Gao, Jianfeng and others},
  journal={Foundations and Trends{\textregistered} in Computer Graphics and Vision},
  volume={16},
  number={1-2},
  pages={1--214},
  year={2024},
  publisher={Now Publishers, Inc.}
}

@misc{o3,
  title        = "{Gpt-o3: A large language model by openai.}",
  author       = "{OpenAI}",
  year         = 2024,
  howpublished = "{\url{https://openai.com/index/hello-gpt-4o/}}"
}

@misc{gemini,
  title        = {Gemini: A family of highly capable multimodal models},
  author       = "{Gemini}",
  year         = 2024,
  howpublished = "{\url{https://gemini.google.com/app}}"
}

@inproceedings{chen2024internvl,
  title={Internvl: Scaling up vision foundation models and aligning for generic visual-linguistic tasks},
  author={Chen, Zhe and Wu, Jiannan and Wang, Wenhai and Su, Weijie and Chen, Guo and Xing, Sen and Zhong, Muyan and Zhang, Qinglong and Zhu, Xizhou and Lu, Lewei and others},
  booktitle=CVPR,
  year={2024}
}

@article{xue2024longvila,
  title={Longvila: Scaling long-context visual language models for long videos},
  author={Xue, Fuzhao and Chen, Yukang and Li, Dacheng and Hu, Qinghao and Zhu, Ligeng and Li, Xiuyu and Fang, Yunhao and Tang, Haotian and Yang, Shang and Liu, Zhijian and others},
  journal={arXiv preprint arXiv:2408.10188},
  year={2024}
}

@misc{lin2023vila,
      title={VILA: On Pre-training for Visual Language Models},
      author={Ji Lin and Hongxu Yin and Wei Ping and Yao Lu and Pavlo Molchanov and Andrew Tao and Huizi Mao and Jan Kautz and Mohammad Shoeybi and Song Han},
      year={2023},
      eprint={2312.07533},
      archivePrefix={arXiv},
      primaryClass={cs.CV}
}

@article{zhang2024longva,
  title={Long Context Transfer from Language to Vision},
  author={Peiyuan Zhang and Kaichen Zhang and Bo Li and Guangtao Zeng and Jingkang Yang and Yuanhan Zhang and Ziyue Wang and Haoran Tan and Chunyuan Li and Ziwei Liu},
  journal={arXiv preprint arXiv:2406.16852},
  year={2024},
  url = {https://arxiv.org/abs/2406.16852}
}

@misc{zhang2024llavanext-video,
  title={LLaVA-NeXT: A Strong Zero-shot Video Understanding Model},
  url={https://llava-vl.github.io/blog/2024-04-30-llava-next-video/},
  author={Zhang, Yuanhan and Li, Bo and Liu, haotian and Lee, Yong jae and Gui, Liangke and Fu, Di and Feng, Jiashi and Liu, Ziwei and Li, Chunyuan},
  month={April},
  year={2024}
}

@article{li2024llava,
  title={Llava-onevision: Easy visual task transfer},
  author={Li, Bo and Zhang, Yuanhan and Guo, Dong and Zhang, Renrui and Li, Feng and Zhang, Hao and Zhang, Kaichen and Zhang, Peiyuan and Li, Yanwei and Liu, Ziwei and others},
  journal={arXiv preprint arXiv:2408.03326},
  year={2024}
}

@article{wang2023chat,
  title={Chat-3d: Data-efficiently tuning large language model for universal dialogue of 3d scenes},
  author={Wang, Zehan and Huang, Haifeng and Zhao, Yang and Zhang, Ziang and Zhao, Zhou},
  journal={arXiv preprint arXiv:2308.08769},
  year={2023}
}

@inproceedings{azuma2022scanqa,
  title={Scanqa: 3d question answering for spatial scene understanding},
  author={Azuma, Daichi and Miyanishi, Taiki and Kurita, Shuhei and Kawanabe, Motoaki},
  booktitle=CVPR,
  year={2022}
}

@inproceedings{ma2022sqa3d,
  title={Sqa3d: Situated question answering in 3d scenes},
  author={Ma, Xiaojian and Yong, Silong and Zheng, Zilong and Li, Qing and Liang, Yitao and Zhu, Song-Chun and Huang, Siyuan},
  booktitle=ICLR,
  year={2022}
}

@inproceedings{jin2023context,
  title={Context-aware alignment and mutual masking for 3d-language pre-training},
  author={Jin, Zhao and Hayat, Munawar and Yang, Yuwei and Guo, Yulan and Lei, Yinjie},
  booktitle=CVPR,
  year={2023}
}

@inproceedings{zhu20233d,
  title={3d-vista: Pre-trained transformer for 3d vision and text alignment},
  author={Zhu, Ziyu and Ma, Xiaojian and Chen, Yixin and Deng, Zhidong and Huang, Siyuan and Li, Qing},
  booktitle=ICCV,
  year={2023}
}

@inproceedings{zhu2024unifying,
  title={Unifying 3d vision-language understanding via promptable queries},
  author={Zhu, Ziyu and Zhang, Zhuofan and Ma, Xiaojian and Niu, Xuesong and Chen, Yixin and Jia, Baoxiong and Deng, Zhidong and Huang, Siyuan and Li, Qing},
  booktitle=ECCV,
  year={2024}
}

@article{yao2024minicpm,
  title={Minicpm-v: A gpt-4v level mllm on your phone},
  author={Yao, Yuan and Yu, Tianyu and Zhang, Ao and Wang, Chongyi and Cui, Junbo and Zhu, Hongji and Cai, Tianchi and Li, Haoyu and Zhao, Weilin and He, Zhihui and others},
  journal={arXiv preprint arXiv:2408.01800},
  year={2024}
}

@article{huang2023embodied,
  title={An embodied generalist agent in 3d world},
  author={Huang, Jiangyong and Yong, Silong and Ma, Xiaojian and Linghu, Xiongkun and Li, Puhao and Wang, Yan and Li, Qing and Zhu, Song-Chun and Jia, Baoxiong and Huang, Siyuan},
  journal={arXiv preprint arXiv:2311.12871},
  year={2023}
}

@article{Robin3D,
    title={Robin3D: Improving 3D Large Language Model via Robust Instruction Tuning}, 
    author={Weitai Kang and Haifeng Huang and Yuzhang Shang and Mubarak Shah and Yan Yan},
    journal={arXiv:2410.00255},
    year={2024},
}

@article{wang2024qwen2,
  title={Qwen2-vl: Enhancing vision-language model's perception of the world at any resolution},
  author={Wang, Peng and Bai, Shuai and Tan, Sinan and Wang, Shijie and Fan, Zhihao and Bai, Jinze and Chen, Keqin and Liu, Xuejing and Wang, Jialin and Ge, Wenbin and others},
  journal={arXiv preprint arXiv:2409.12191},
  year={2024}
}

@inproceedings{achlioptas2020referit3d,
  title={Referit3d: Neural listeners for fine-grained 3d object identification in real-world scenes},
  author={Achlioptas, Panos and Abdelreheem, Ahmed and Xia, Fei and Elhoseiny, Mohamed and Guibas, Leonidas},
  booktitle={Computer Vision--ECCV 2020: 16th European Conference, Glasgow, UK, August 23--28, 2020, Proceedings, Part I 16},
  pages={422--440},
  year={2020},
  organization={Springer}
}

@inproceedings{sisbot2007spatial,
  title={Spatial reasoning for human robot interaction},
  author={Sisbot, Emrah Akin and Marin, Luis F and Alami, Rachid},
  booktitle={2007 IEEE/RSJ International Conference on Intelligent Robots and Systems},
  pages={2281--2287},
  year={2007},
  organization={IEEE}
}

@article{landsiedel2017review,
  title={A review of spatial reasoning and interaction for real-world robotics},
  author={Landsiedel, Christian and Rieser, Verena and Walter, Matthew and Wollherr, Dirk},
  journal={Advanced Robotics},
  volume={31},
  number={5},
  pages={222--242},
  year={2017},
  publisher={Taylor \& Francis}
}

@inproceedings{marza2022teaching,
  title={Teaching agents how to map: Spatial reasoning for multi-object navigation},
  author={Marza, Pierre and Matignon, Laetitia and Simonin, Olivier and Wolf, Christian},
  booktitle={2022 IEEE/RSJ International Conference on Intelligent Robots and Systems (IROS)},
  pages={1725--1732},
  year={2022},
  organization={IEEE}
}

@inproceedings{chen2021scan2cap,
  title={Scan2cap: Context-aware dense captioning in rgb-d scans},
  author={Chen, Zhenyu and Gholami, Ali and Nie{\ss}ner, Matthias and Chang, Angel X},
  booktitle={Proceedings of the IEEE/CVF conference on computer vision and pattern recognition},
  pages={3193--3203},
  year={2021}
}

@inproceedings{chen2020scanrefer,
  title={Scanrefer: 3d object localization in rgb-d scans using natural language},
  author={Chen, Dave Zhenyu and Chang, Angel X and Nie{\ss}ner, Matthias},
  booktitle={European conference on computer vision},
  pages={202--221},
  year={2020},
  organization={Springer}
}

@inproceedings{Multi3DRefer,
  title={Multi3DRefer: Grounding Text Description to Multiple 3D Objects},
  author={Zhang, Yiming and Gong, ZeMing and Chang, Angel X},
  booktitle=ICCV,
  year={2023}
}

@article{dai2017bundlefusion,
  title={Bundlefusion: Real-time globally consistent 3d reconstruction using on-the-fly surface reintegration},
  author={Dai, Angela and Nie{\ss}ner, Matthias and Zollh{\"o}fer, Michael and Izadi, Shahram and Theobalt, Christian},
  journal={ACM Transactions on Graphics (ToG)},
  volume={36},
  number={4},
  pages={1},
  year={2017},
  publisher={ACM New York, NY, USA}
}

@article{chen2021d3net,
  title={D3net: A speaker-listener architecture for semi-supervised dense captioning and visual grounding in rgb-d scans},
  author={Chen, Dave Zhenyu and Wu, Qirui and Nie{\ss}ner, Matthias and Chang, Angel X},
  year={2021}
}

@inproceedings{zhao20213dvg,
  title={3dvg-transformer: Relation modeling for visual grounding on point clouds},
  author={Zhao, Lichen and Cai, Daigang and Sheng, Lu and Xu, Dong},
  booktitle={Proceedings of the IEEE/CVF International Conference on Computer Vision},
  pages={2928--2937},
  year={2021}
}

@inproceedings{schult2023mask3d,
  title={Mask3d: Mask transformer for 3d semantic instance segmentation},
  author={Schult, Jonas and Engelmann, Francis and Hermans, Alexander and Litany, Or and Tang, Siyu and Leibe, Bastian},
  booktitle={2023 IEEE International Conference on Robotics and Automation (ICRA)},
  year={2023},
  organization={IEEE}
}

@article{kolodiazhnyi2024unidet3d,
  title={UniDet3D: Multi-dataset Indoor 3D Object Detection},
  author={Kolodiazhnyi, Maksim and Vorontsova, Anna and Skripkin, Matvey and Rukhovich, Danila and Konushin, Anton},
  journal={arXiv preprint arXiv:2409.04234},
  year={2024}
}

@inproceedings{dai2017scannet,
  title={Scannet: Richly-annotated 3d reconstructions of indoor scenes},
  author={Dai, Angela and Chang, Angel X and Savva, Manolis and Halber, Maciej and Funkhouser, Thomas and Nie{\ss}ner, Matthias},
  booktitle=CVPR,
  year={2017}
}

@article{baruch2021arkitscenes,
  title={Arkitscenes: A diverse real-world dataset for 3d indoor scene understanding using mobile rgb-d data},
  author={Baruch, Gilad and Chen, Zhuoyuan and Dehghan, Afshin and Dimry, Tal and Feigin, Yuri and Fu, Peter and Gebauer, Thomas and Joffe, Brandon and Kurz, Daniel and Schwartz, Arik and others},
  journal={arXiv preprint arXiv:2111.08897},
  year={2021}
}

@inproceedings{yeshwanth2023scannet++,
  title={Scannet++: A high-fidelity dataset of 3d indoor scenes},
  author={Yeshwanth, Chandan and Liu, Yueh-Cheng and Nie{\ss}ner, Matthias and Dai, Angela},
  booktitle={Proceedings of the IEEE/CVF International Conference on Computer Vision},
  pages={12--22},
  year={2023}
}

@book{nadel2008hippocampus,
  author    = {Nadel, Lynn},
  title     = {The Hippocampus and Context Revisited},
  year      = {2008},
  publisher = {Oxford University Press}
}

@article{tolman1948cognitive,
  author    = {Tolman, E. C.},
  title     = {Cognitive maps in rats and men},
  journal   = {Psychological Review},
  year      = {1948},
  volume    = {55},
  number    = {4},
  pages     = {189--208}
}

@inproceedings{gu2022vision,
  title={Vision-and-Language Navigation: A Survey of Tasks, Methods, and Future Directions},
  author={Gu, Jing and Stefani, Eliana and Wu, Qi and Thomason, Jesse and Wang, Xin},
  booktitle={Proceedings of the 60th Annual Meeting of the Association for Computational Linguistics (Volume 1: Long Papers)},
  pages={7606--7623},
  year={2022}
}

@article{ma2025spatialreasoner,
  title={SpatialReasoner: Towards Explicit and Generalizable 3D Spatial Reasoning},
  author={Ma, Wufei and Chou, Yu-Cheng and Liu, Qihao and Wang, Xingrui and de Melo, Celso and Chen, Jieneng and Xie, Jianwen and Yuille, Alan},
  journal={arXiv preprint arXiv:2504.20024},
  year={2025}
}

@article{liao2025improved,
  title={Improved visual-spatial reasoning via r1-zero-like training},
  author={Liao, Zhenyi and Xie, Qingsong and Zhang, Yanhao and Kong, Zijian and Lu, Haonan and Yang, Zhenyu and Deng, Zhijie},
  journal={arXiv preprint arXiv:2504.00883},
  year={2025}
}

@article{li2023m3dbench,
  title={M3DBench: Let's Instruct Large Models with Multi-modal 3D Prompts},
  author={Li, Mingsheng and Chen, Xin and Zhang, Chi and Chen, Sijin and Zhu, Hongyuan and Yin, Fukun and Yu, Gang and Chen, Tao},
  journal={arXiv preprint arXiv:2312.10763},
  year={2023}
}

@inproceedings{li20243dmit,
  title={3dmit: 3d multi-modal instruction tuning for scene understanding},
  author={Li, Zeju and Zhang, Chao and Wang, Xiaoyan and Ren, Ruilong and Xu, Yifan and Ma, Ruifei and Liu, Xiangde and Wei, Rong},
  booktitle={2024 IEEE International Conference on Multimedia and Expo Workshops (ICMEW)},
  pages={1--5},
  year={2024},
  organization={IEEE}
}

@inproceedings{liu2024grounding,
  title={Grounding dino: Marrying dino with grounded pre-training for open-set object detection},
  author={Liu, Shilong and Zeng, Zhaoyang and Ren, Tianhe and Li, Feng and Zhang, Hao and Yang, Jie and Jiang, Qing and Li, Chunyuan and Yang, Jianwei and Su, Hang and others},
  booktitle={European Conference on Computer Vision},
  pages={38--55},
  year={2024},
  organization={Springer}
}

@inproceedings{kirillov2023segment,
  title={Segment anything},
  author={Kirillov, Alexander and Mintun, Eric and Ravi, Nikhila and Mao, Hanzi and Rolland, Chloe and Gustafson, Laura and Xiao, Tete and Whitehead, Spencer and Berg, Alexander C and Lo, Wan-Yen and others},
  booktitle={Proceedings of the IEEE/CVF international conference on computer vision},
  pages={4015--4026},
  year={2023}
}

@article{yang2025magma,
  title={Magma: A foundation model for multimodal ai agents},
  author={Yang, Jianwei and Tan, Reuben and Wu, Qianhui and Zheng, Ruijie and Peng, Baolin and Liang, Yongyuan and Gu, Yu and Cai, Mu and Ye, Seonghyeon and Jang, Joel and others},
  journal={arXiv preprint arXiv:2502.13130},
  year={2025}
}

@article{zhu2024towards,
  title={Towards flexible visual relationship segmentation},
  author={Zhu, Fangrui and Yang, Jianwei and Jiang, Huaizu},
  journal={Advances in Neural Information Processing Systems},
  volume={37},
  pages={107633--107661},
  year={2024}
}

@article{gu2025phyworldbench,
  title={" PhyWorldBench": A Comprehensive Evaluation of Physical Realism in Text-to-Video Models},
  author={Gu, Jing and Liu, Xian and Zeng, Yu and Nagarajan, Ashwin and Zhu, Fangrui and Hong, Daniel and Fan, Yue and Yan, Qianqi and Zhou, Kaiwen and Liu, Ming-Yu and others},
  journal={arXiv preprint arXiv:2507.13428},
  year={2025}
}

\clearpage
\appendix

\section{Details of \model{} Prompting Strategy.}
\begin{figure}[t]
    \centering
    \includegraphics[width=\linewidth]{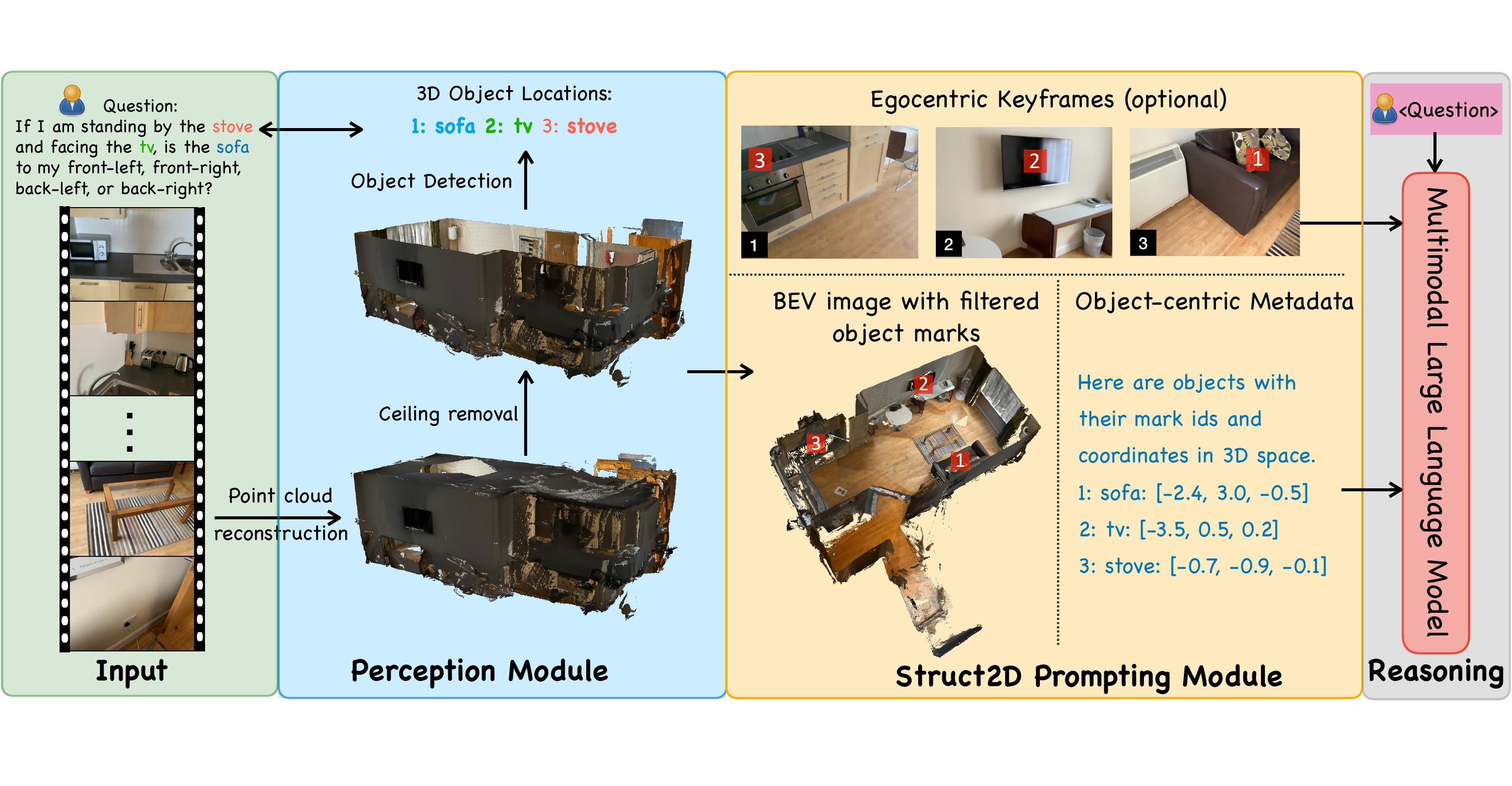}
    \caption{\label{fig:prompt_structure} \textbf{Overview of the \model{} Prompting Framework.} Given an egocentric video and a spatial question, we first reconstruct a 3D point cloud and remove the ceiling for a clear top-down view. Objects are detected in 3D space, and a bird’s-eye-view (BEV) image is rendered with object marks projected onto the floor plane. These object marks are filtered based on the content of the question. We also extract egocentric keyframes by projecting 3D bounding box centers onto the video, when appearance cues are needed. Object-centric metadata—including object IDs and 3D coordinates—is encoded as text. The structured 2D visual and textual inputs are then fed into a multimodal large language model for spatial reasoning. }
    
\end{figure}

Figure~\ref{fig:prompt_structure} illustrates the overall \model{} prompting framework, which transforms egocentric 3D scene input into structured 2D representations for spatial reasoning. Given an input video and a spatial question, we first reconstruct a 3D point cloud from RGB-D frames and remove the ceiling to obtain a clear top-down view of the scene. Object detection is then performed in 3D space, and detected objects are projected onto a bird’s-eye-view (BEV) image to produce a layout of the environment. These object marks are filtered to retain only those relevant to the input question.

We optionally extract egocentric keyframes to capture detailed object appearances. Keyframes are selected by projecting 3D object bounding boxes onto sampled video frames and depth maps, and identifying views where each object is both visible and unobstructed. Object-centric metadata—including object categories and 3D coordinates—is encoded as text and used as part of the prompt input.

Algorithm~\ref{alg:ours} outlines the core procedure for constructing the \model{} prompt. Given an input video \(\mathbf{V}\), depth frames \(\mathbf{D}\), a reconstructed 3D scene \(\mathcal{P}\), and a set of target objects \(\mathcal{O}\), we begin by rendering a BEV image \(v\) and projecting each object \(o_i \in \mathcal{O}\) into the view using the RGB camera parameters \(C_{\text{rgb}}\). The 2D projections are then drawn as object marks on the image.

To select keyframes, we sample \(N\) RGB-D frames and iteratively check for visibility of objects not yet covered in the BEV. For each candidate frame \(I_i\), we project the remaining unseen objects onto both the frame and its depth map. If a valid projection exists (i.e., the projected location lies within the image and has valid depth), the object mark is rendered and the frame is added to the keyframe set \(\mathcal{I}_{\text{keys}}\). This process continues until all relevant objects are covered. The final prompt consists of \ding{182} a BEV image with filtered marks, \ding{183} optional keyframes containing visible objects, and \ding{184} object metadata text, all of which are passed to a multimodal large language model for reasoning.

This framework allows the MLLM to perform 3D spatial reasoning from 2D visual and textual inputs, without requiring direct access to raw 3D data at inference time. It enables scalable, flexible spatial understanding grounded in realistic perception outputs.

\begin{algorithm}[ht]
\caption{{\model} Visual Prompting}
\label{alg:ours}
\begin{algorithmic}[1]
\Require Input video $\mathbf{V}$, Depth frames $\mathbf{D}$, Reconstructed 3D scene $\mathcal{P}$, Objects of interest $\mathcal{O}$, RGB camera parameters $\mathbf{C}_{\text{rgb}}$, Depth camera parameters $\mathbf{C}_{\text{d}}$
\State Render a Bird's Eye View image: $\mathbf{v} \gets \text{BEV}(\mathcal{P})$
\For{$o_i \in \mathcal{O}$}
    \State Project $o_i$ onto $\mathbf{v}$: $p_i \gets \text{Project}(o_i, \mathbf{v}, \mathbf{C}_{\text{rgb}})$
    \State Update view: $\mathbf{v} \gets \text{Add-Mark}(\mathbf{v}, p_i)$
\EndFor
\State Sample $N$ frames: $\mathbf{I}, \mathbf{D_I} \gets \text{Sample}(\mathbf{V}, \mathbf{D})$
\State Initialize key frame set: $\mathbf{I}_\text{keys} \gets \{\}$
\State Initialize found objects set: $\mathcal{O}_\text{F} \gets \{\}$ 
\For{$\mathbf{I_i} \in \mathbf{I}$ and $\mathbf{D_i}\in\mathbf{D_I}$}
    \State $b_i \gets \text{False}$
    \For{$o_j \in \mathcal{O}$ and $\notin\mathcal{O}_\text{F}$}
        \State Project $o_j$ onto $\mathbf{I_i}$ and $\mathbf{D_i}$: $p_j^I \gets \text{Project}(o_j, \mathbf{I_i}, \mathbf{C}_{\text{rgb}})$, $p_j^D \gets \text{Project}(o_j, \mathbf{D_i}, \mathbf{C}_{\text{d}})$
        \If{$p_j^I \in \mathbf{I_i}$ and $p_j^D \in \mathbf{D_i}$ and $p_j^D \ge0$}
            \State $b_i \gets \text{True}$
            \State Update frame: $\mathbf{I_i}\gets\text{Add-Mark}(\mathbf{I_i}, p_j^I)$
            \State Add object to set: $\mathcal{O}_\text{F} \gets \mathcal{O}_\text{F}\cup\{o_i\}$
        \EndIf
    \EndFor
    \If{$b_i$}
        \State Add to key frame set: $\mathbf{I}_\text{keys} \gets \mathbf{I}_\text{keys} \cup \{\mathbf{I_i}\}$
    \EndIf
\EndFor
\Ensure Informative BEV view $\mathbf{v}$ and key frame set $\mathbf{I}_\text{keys}$
\end{algorithmic}
\end{algorithm}

\begin{figure}
    \centering
    \includegraphics[width=\linewidth]{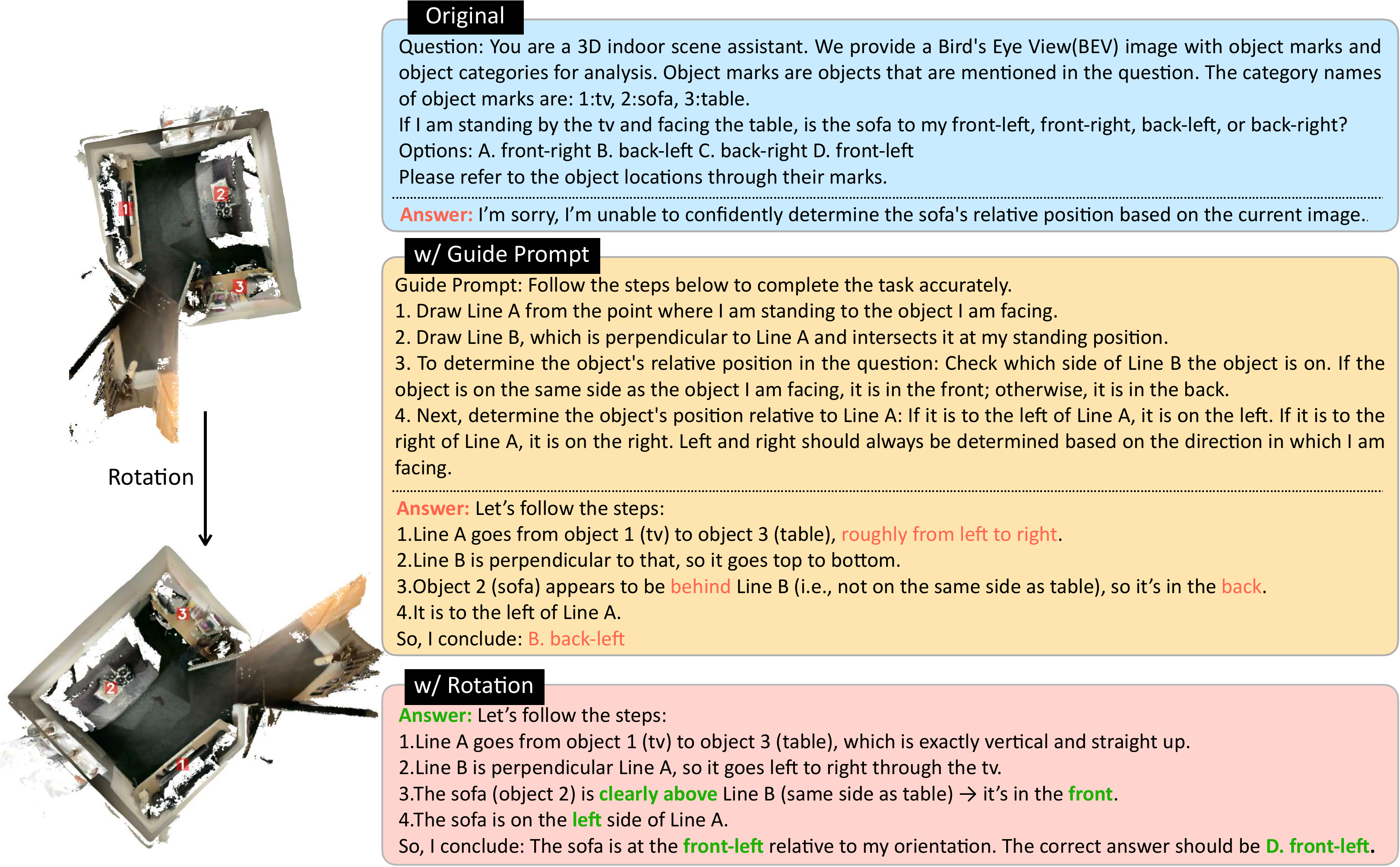}
    \caption{\label{fig:zero-shot-vis} \textbf{Effects of guide prompting and rotation correction in relative spatial reasoning of the zero-shot prompting.} We show different variations of our \model{}. The original model struggles to interpret spatial relationships using the raw BEV image and object marks alone, failing to answer correctly. With the addition of a structured guide prompt, the model is able to decompose the reasoning process into geometric steps, though it still arrives at an incorrect answer. When object orientation is further corrected via rotation alignment, the spatial reference frame becomes more intuitive, and the model's reasoning becomes clearer and more accurate. This demonstrates the importance of both prompting design and canonical alignment in unlocking MLLMs' spatial understanding abilities. (\textcolor{red}{Red} texts are wrong answers; \textcolor{green}{Green} texts are correct ones.)}
    
\end{figure}

\begin{figure}
    \centering
    \includegraphics[width=\linewidth]{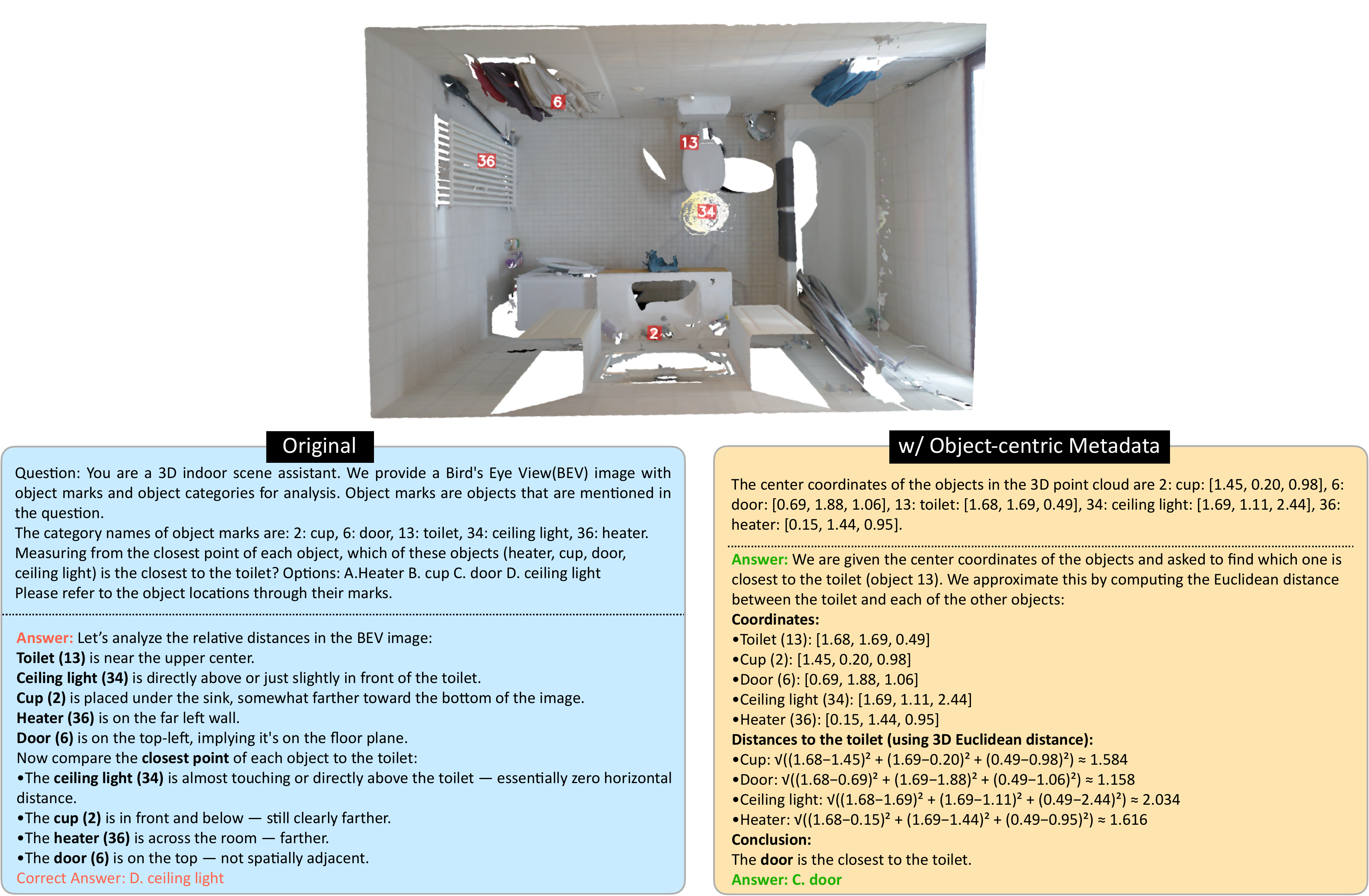}
    \caption{\label{fig:zero-shot-vis-2}\textbf{Effect of object-centric metadata for precise spatial reasoning.} Originally, the model attempts to estimate distances based solely on the spatial layout in the BEV image but fails to identify the correct object closest to the toilet. In contrast, with access to object-centric metadata—specifically, 3D coordinates of each object—the model can compute accurate Euclidean distances and correctly identify the nearest object. This example highlights how structured metadata enhances geometric reasoning and helps avoid ambiguity in visual interpretation.
(\textcolor{red}{Red} text indicates incorrect reasoning; \textcolor{green}{Green} text indicates the correct answer.)
}
    
\end{figure}

\noindent\textbf{Qualitative comparison of our \model{} prompting.}
To better understand the impact of prompt design on spatial reasoning, we conduct qualitative analyses highlighting two key aspects of our framework: reasoning guidance, object orientation, and structured metadata. As shown in Figure~\ref{fig:zero-shot-vis}, when the model is prompted only with a BEV image and object marks, it struggles to accurately resolve relative spatial relationships. Adding a structured guide prompt enables the model to decompose the task into interpretable geometric steps, though it may still fail without an aligned reference frame. Once the BEV is rotation-aligned with the agent's viewpoint, the reasoning becomes more intuitive, leading to the correct answer. Similarly, in Figure~\ref{fig:zero-shot-vis-2}, we illustrate the benefit of object-centric metadata. Without access to precise coordinates, the model must estimate distances visually, which can lead to errors. When provided with 3D object positions, the model can directly compute spatial relations such as Euclidean distances, significantly improving its accuracy on localization tasks. These examples highlight how prompt structure---through guided reasoning and geometric priors---plays a crucial role in unlocking spatial understanding in MLLMs.

\section{Details of \data{}}
\noindent \textbf{Overview.}
\data{} is a large-scale instruction tuning dataset aimed at enabling spatial reasoning and scene understanding in indoor 3D environments using only 2D projected inputs. It contains over 200K question-answer (QA) pairs derived from 6K richly annotated indoor scenes drawn from ScanNet~\cite{dai2017scannet}, ScanNet++\cite{yeshwanth2023scannet++}, and ARKitScenes\cite{baruch2021arkitscenes}. Each QA instance is paired with structured scene- and object-level metadata, allowing models to learn spatial concepts without relying on explicit 3D feature representations during training.

The dataset spans eight categories of spatial reasoning tasks, such as object attribute identification, relative localization, and egocentric navigation. Each QA pair follows an instruction-style format and includes:
\begin{itemize}[nosep,leftmargin=2em]
\item A natural language question,
\item A concise short-form answer,
\item A long-form answer, when applicable, containing step-by-step reasoning or contextual elaboration,
\item Accompanying metadata including relevant object marks, spatial coordinates, and references to visual input modalities (\eg, BEV image, selected keyframes).
\end{itemize}
Long-form answers are provided selectively for tasks that benefit from explicit reasoning or contextual understanding. For categories requiring direct factual responses—such as object counting or binary verification—only short-form answers are used. This balanced design ensures effective supervision across tasks of varying complexity, while maintaining interpretability and richness in reasoning. We next describe the construction process for each task category in detail.

\noindent\textbf{Object counting.}To construct object counting questions, we begin by sampling a scene from the training split of the source datasets and extracting its ground-truth object annotations. A target object category (\eg, \textit{chair}) is randomly selected from the annotated instances within the scene. A QA pair is then generated using a templated prompt such as \textit{``How many {class label}(s) are there in this room?''}, paired with the correct numerical count as the answer. To improve linguistic diversity and fluency, we further augment these questions by prompting ChatGPT to generate alternative phrasings with equivalent semantic meaning.

\textbf{Spatial Relationship.}
This category evaluates a model’s ability to reason about the directional relationships between objects in a 3D scene from an egocentric perspective. Following the formulation in VSI-Bench~\cite{yang2024thinking}, we focus on the subtask of \textit{relative direction}, where the goal is to identify the directional location of a target object based on a specified standing point and facing direction.

To construct each QA pair, we begin by computing the 3D centers of all objects in the scene and projecting them onto the 2D BEV image. We then sample a triplet of objects representing the roles of \textit{standing}, \textit{facing}, and \textit{target}, while filtering out ambiguous categories (\eg, object clusters or large connected instances) and enforcing a minimum pixel-distance threshold to ensure spatial distinguishability. The <\textit{standing}, \textit{facing}> vector defines the forward direction of the agent, and the <\textit{standing}, \textit{target}> vector is used to determine the relative orientation of the target object. The angular offset between these vectors is then discretized into directional bins such as \textit{front-left}, \textit{right}, or \textit{back}, producing the correct label.

We format each QA pair using a natural language template (\eg, \textit{``If I am standing by the TV and facing the refrigerator, is the sink to my left, right, or back?''}) and provide the short-form directional answer. To enhance both linguistic variation and model supervision, we further augment each instance using ChatGPT, which paraphrases the question and generates a long-form answer that walks through the step-by-step reasoning process under the egocentric frame of reference.

\textbf{Comparative Reasoning.}
This category involves tasks where the model must compare spatial attributes among multiple objects. We focus on \textit{relative distance comparison}, where the objective is to identify which candidate object is closest or farthest from a given reference object.

To construct such questions, we first select a reference object whose identity is unambiguous based on its class label. Next, we sample a set of candidate objects, including multiple instances—potentially of the same class—to encourage instance-level discrimination. In contrast to reference selection, we do not filter ambiguous or repeated categories among the candidates, as the goal is to challenge the model to reason over instance-specific spatial relations.

We compute the 3D centroid of each object using the center of its oriented bounding box and measure pairwise Euclidean distances between the reference and each candidate. Based on the ranking of these distances, we generate a templated question, such as \textit{``Measuring from the closest point of each object, which of these objects ({candidate labels}) is closest to the {reference object}?''}, along with the correct answer derived from the computed rankings.

To enhance linguistic variation and encourage deeper reasoning, we further augment each instance using ChatGPT, which paraphrases the question and generates a long-form answer. These enriched responses guide the model through comparative spatial reasoning before producing the final answer.

\textbf{Quantitative Spatial Measuring.}
This category targets tasks requiring the model to reason about metric properties in 3D space, such as object size, spatial extent, and inter-object distance. We focus on the \textit{object absolute distance} subtask, where the model needs to estimate the physical distance between two specified objects within a scene.

To construct these questions, we begin by selecting two distinct objects with clearly identifiable class labels to avoid semantic ambiguity. Using the oriented bounding box annotations, we extract all eight corner points for each object and compute the minimum Euclidean distance across all point pairs—this serves as the ground-truth physical distance between the two objects. Based on this calculation, we generate templated questions such as: \textit{``Measuring from the closest point of each object, what is the distance between the {object1} and the {object2} (in meters)?''}

To enhance supervision and promote reasoning transparency, we further use ChatGPT to produce long-form answers. These responses walk through the spatial computation process, prompting the model to conceptually simulate pairwise distance comparisons before arriving at the correct numerical answer.

\textbf{Egocentric Navigation.}
This category focuses on tasks that require the model to plan navigation routes from an egocentric perspective, reasoning about object references, turning actions, and scene layout. The goal is to simulate how an embodied agent would traverse a 3D space by following instructions grounded in object-level references.

To construct these tasks, we first sample up to 15 candidate objects per scene and project their 3D centers onto the BEV image. Each object is visually marked in the BEV, and a mark-to-label dictionary string is generated to facilitate object identification. These scene representations are then passed to ChatGPT to generate plausible navigation routes in natural language.

Route generation is guided by several constraints:
\ding{182} Each route must consist of a sequence of consecutive object marks (IDs) that an agent can follow.
\ding{183} At each step, the agent must perform a local navigation action (e.g., turn left, turn right, pass by).
\ding{184} Routes must avoid collisions with irrelevant or obstructing objects.
\ding{185} Each path should span 3 to 5 objects to ensure sufficient reasoning complexity.

All generated routes undergo human review to ensure spatial plausibility. Invalid routes are discarded, and valid ones are further augmented via route reversal and sub-segmentation to increase diversity.

To determine the action sequence along the path, we randomly choose a facing object at the starting point to establish the initial egocentric orientation. For each transition between objects, we compute the vector from the current object to the next and compare it with the current facing direction to infer the correct action (e.g., turn left, go forward). These navigation actions form the short-form answer.

For each object along the route, we apply our keyframe selection algorithm to extract egocentric views from the original video. These keyframes, combined with the object labels, are used to prompt ChatGPT to generate rich textual descriptions of each waypoint. Finally, we instruct ChatGPT to produce long-form answers that walk through the full navigation route, step by step, reasoning over orientation shifts and identifying the appropriate navigation action at each stage.

\textbf{Other Categories.}
The remaining task types—such as object attribute identification and binary attribute verification—are constructed by augmenting QA pairs from existing 3D vision-language datasets, including ScanQA~\cite{azuma2022scanqa}, SQA3D~\cite{ma2022sqa3d}, Scan2Cap~\cite{chen2021scan2cap}, ScanRefer~\cite{chen2020scanrefer}, and Multi3DRefer~\cite{Multi3DRefer}. These datasets provide scene-specific questions grounded in the ScanNet environment and collectively cover all eight spatial reasoning categories defined in \data{}.

To adapt these examples for instruction tuning, we first use ChatGPT to rephrase each question into a more natural and instructional style. For tasks requiring reasoning, we also prompt ChatGPT to generate long-form answers that walk through the inference process. For visual grounding, we localize referenced objects using two approaches: when object IDs are available, we apply our keyframe detection method to extract representative egocentric views. In datasets with descriptive references (e.g., SQA3D), we extract referring expressions with ChatGPT and apply Grounding DINO~\cite{liu2024grounding} and SAM~\cite{kirillov2023segment} to identify and segment the mentioned objects.

The resulting keyframes are paired with each QA instance to serve as visual inputs during fine-tuning. This pipeline enables instruction tuning on complex, object-centric spatial tasks while relying only on 2D visual projections and avoiding the need for explicit 3D geometry at training time.

\begin{figure}
    \centering
    \includegraphics[width=\linewidth]{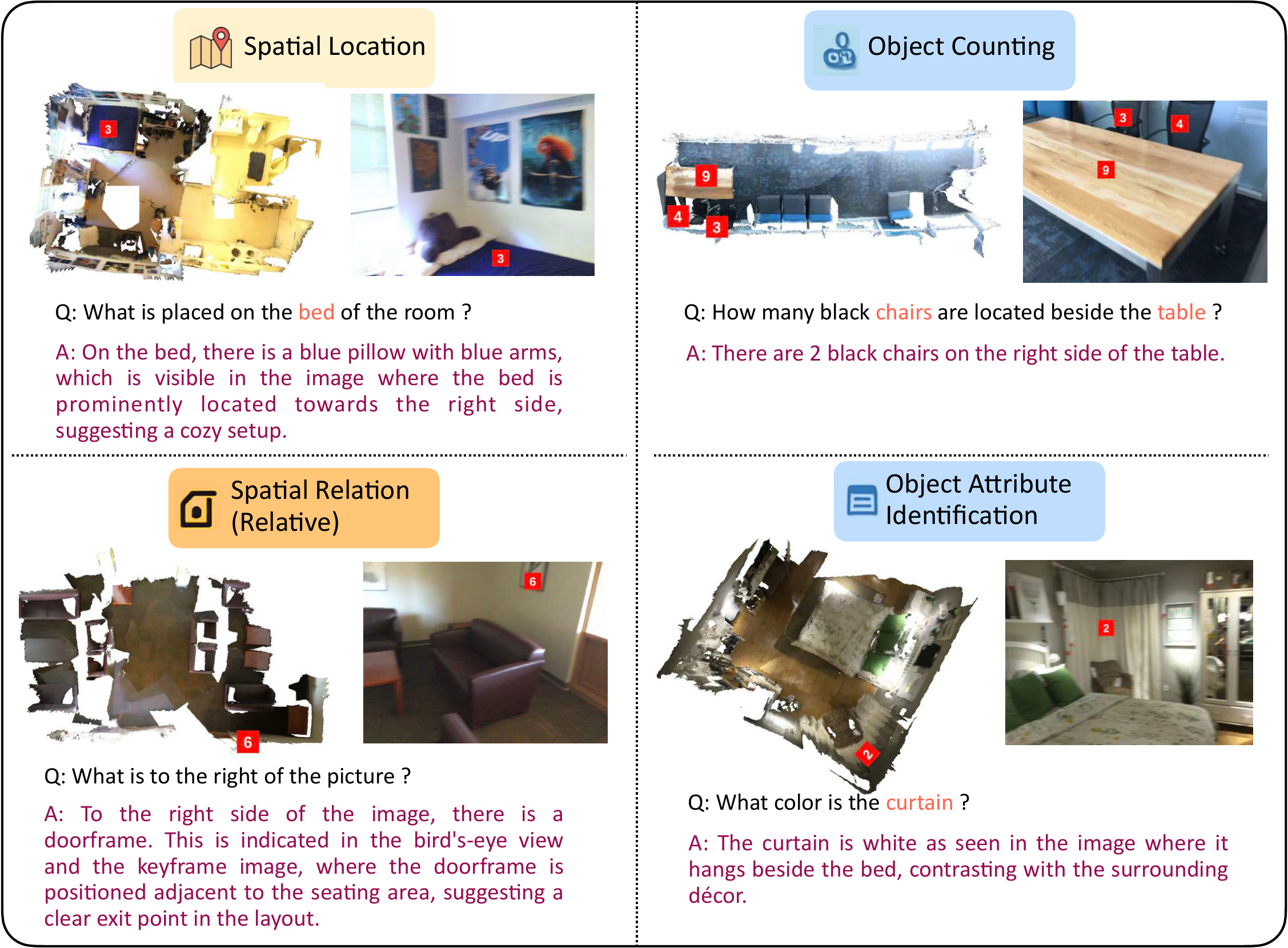}
    \caption{\label{fig:train_qa_1}\textbf{Examples of QA pairs from \data{} used for instruction tuning.} Each example illustrates a distinct category of spatial reasoning: spatial localization, object counting, spatial relationship, and object attribute identification. For each question, the model is provided with a BEV image annotated with object marks, and optionally an egocentric keyframe to enhance visual grounding. The answers include descriptive reasoning grounded in object positions and appearances, enabling the model to learn to associate structured 2D inputs with fine-grained spatial understanding.}
    
\end{figure}

\begin{figure}
    \centering
    \includegraphics[width=\linewidth]{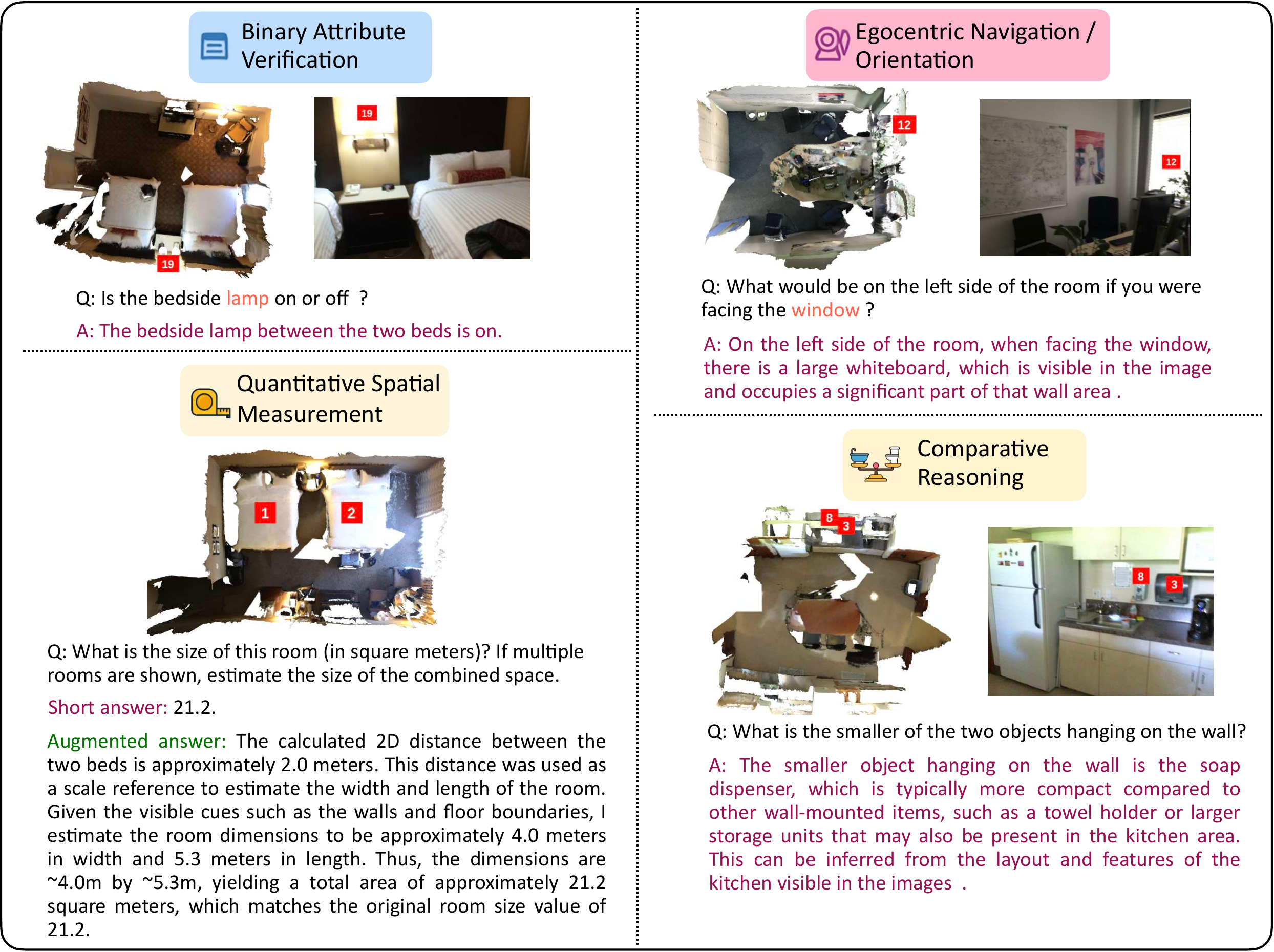}
    \caption{\label{fig:train_qa_2} \textbf{Additional QA examples from \data{} covering diverse spatial reasoning categories.} This figure showcases examples from the remaining categories in our dataset: binary attribute verification, egocentric navigation and orientation, quantitative spatial measurement, and comparative reasoning. Each QA pair is grounded in structured 2D visual inputs (BEV views and keyframes) and enriched with object marks and contextual metadata. These examples demonstrate the model’s ability to reason about object states, egocentric spatial references, metric estimations, and relative comparisons—key competencies for embodied spatial understanding.}

\end{figure}

\section{Implementation Details}
We use Qwen2.5VL~\cite{wang2024qwen2} as the base multimodal large language model (MLLM) for instruction tuning. During training, the model receives BEV images with filtered object marks and object-centric metadata as core inputs. For tasks requiring visual cues such as object color or quantity, we additionally provide egocentric keyframes. The BEV images are resized to $640 \times 640$, with object marks adaptively scaled based on their original image resolution. Keyframes are resized to $256 \times 246$ and stitched into compact 1×2 or 2×4 grids, enabling efficient batch loading and reducing GPU memory consumption.

To support reasoning supervision, we adopt a task-specific output format. For complex spatial reasoning tasks—such as relative direction estimation or route planning—we wrap the reasoning process between special tokens \texttt{<think>} and \texttt{</think>}, followed by the final answer enclosed within \texttt{<answer>} and \texttt{</answer>}. For tasks focused on appearance or simple measurements, the model is trained to generate direct short-form answers without explicit reasoning traces.
The model is trained for one epoch using a base learning rate of 2e-6 with cosine annealing, taking approximately 8 hours on 8×H200 GPUs.

At evaluation time, we follow standard practices from prior work~\cite{huang2023chat,qi2025gpt4scene}, reconstructing point clouds offline using BundleFusion~\cite{dai2017bundlefusion}, detecting 3D objects using Mask3D\cite{schult2023mask3d} and UniDet~\cite{kolodiazhnyi2024unidet3d}, and projecting the results to produce BEV images and 2D object marks. For object-level grounding, we apply a rule-based method to identify the relevant objects mentioned in each question.

\section{Results on 3D Grounding and 3D Dense Captioning}

\begin{table*}[t]
\centering
\caption{\textbf{3D Grounding Evaluation on ScanRefer~\cite{chen2020scanrefer} and Multi3DRefer~\cite{Multi3DRefer} datasets.}}
\label{tab:grounding}
\begin{tabular}{lcccc}
    \toprule
    \multirow{2}{*}{\textbf{Methods}} & \multicolumn{2}{c}{\textbf{ScanRefer (val)}} & \multicolumn{2}{c}{\textbf{Multi3DRefer (val)}} \\
    \cmidrule(lr){2-3} \cmidrule(lr){4-5}
     & Acc@0.25 & Acc@0.50 & all F1@0.25 & all F1@0.50 \\
    \midrule
    \rowA \multicolumn{5}{l}{\textit{Task-Specific Model}} \vspace{1mm} \\
    
    3DVG-Transformer~\cite{zhao20213dvg} & 47.6 & 34.7 & -- & 25.5 \\
    
    3DJCG~\cite{3djcg} & 49.6 & 37.3 & -- & 26.6 \\
    
    D3Net~\cite{chen2021d3net}  & -- & 37.9 & -- & 32.2 \\
    
    M3DRef-CLIP~\cite{Multi3DRefer} & 51.9 & 44.7 & 42.8 & 38.4 \\
    
    \midrule
    \rowA \multicolumn{5}{l}{\textit{3D LLM Based Model}} \vspace{1mm} \\
    
    Chat-Scene~\cite{huang2023chat} & 55.5 & 50.2 & 57.1 & 52.4 \\
    
    \midrule
    \rowA \multicolumn{5}{l}{\textit{Vision LLM Based Model}} \vspace{1mm} \\
    
    Qwen2-VL-7B~\cite{wang2024qwen2} & 5.4 & 5.1 & 21.1 & 19.9 \\
    
    Qwen2-VL-7B (GPT4Scene~\cite{qi2025gpt4scene})  & 40.5 & 36.7 & 45.4& 42.1\\
    Qwen2.5-VL-7B (Ours) & 51.7 & 48.5 & 42.1 & 40.6 \\
    \bottomrule
\end{tabular}
\end{table*}

\begin{table*}[t]
\centering
\caption{\textbf{3D Dense Captioning Evaluation on Scan2Cap~\cite{chen2021scan2cap} dataset.}}
\label{tab:dense_caption}
\begin{tabular}{lcccc}
    \toprule
    \multirow{2}{*}{\textbf{Methods}} & \multicolumn{2}{c}{IoU@0.25} & \multicolumn{2}{c}{IoU@0.5} \\
    \cmidrule(lr){2-3} \cmidrule(lr){4-5}
    & BLEU-4 & ROUGE & BLEU-4 & ROUGE \\
    
    \midrule
    \rowA \multicolumn{5}{l}{\textit{Task-Specific Model}} \vspace{1mm} \\
    
    Scan2Cap~\cite{chen2021scan2cap} & 34.2 & 55.3 & 23.3 & 44.5 \\
    
    3DJCG~\cite{3djcg} & 40.2 & 59.2 & 31.0 & 50.8 \\
    
    X-Trans2Cap~\cite{X-trans2cap} & 35.7 & 54.7 & 25.1 & 45.3 \\
    
    3D-VisTA~\cite{zhu20233d} & 36.5 & 57.6 & 34.0 & 54.3 \\
    
    Vote2Cap-DETR~\cite{vote2cap-detr} & 39.3 & 59.3 & 34.5 & 54.4 \\
    
    \midrule
    \rowA \multicolumn{5}{l}{\textit{3D LLM Based Model}} \vspace{1mm} \\
    
    LL3DA~\cite{chen2024ll3da} & 41.4 & 59.5 & 36.8 & 55.1 \\
    
    LEO~\cite{huang2023embodied} & -- & -- & 36.9 & 57.8 \\
    
    Chat-Scene~\cite{huang2023chat} & 38.2 & 60.6 & 36.3 & 58.1 \\
    
    Robin3D~\cite{Robin3D} & -- & -- & 38.4 & -- \\
    
    \midrule
    \rowA \multicolumn{5}{l}{\textit{Vision LLM Based Model}} \vspace{1mm} \\
    
    Qwen2-VL-7B~\cite{wang2024qwen2} & 3.8 & 24.7 & 3.8 & 24.6\\
    
    Qwen2-VL-7B (GPT4Scene~\cite{qi2025gpt4scene})  & 36.3  & 57.6 & 34.2 & 55.2 \\
    
    Qwen2.5-VL-7B (Ours) & 34.8 & 57.0 & 32.7 & 54.5 \\
    
    \bottomrule
\end{tabular}
\end{table*}

\noindent \textbf{Quantitative results.}
Tables~\ref{tab:grounding} and \ref{tab:dense_caption} present our model’s performance on 3D grounding (ScanRefer, Multi3DRefer) and dense captioning (Scan2Cap) benchmarks. While our method does not achieve the highest scores under rule-based metrics such as all F1@0.25/0.5 and BLEU/ROUGE, it consistently delivers competitive results compared to existing vision-language baselines. Importantly, our approach does not rely on point cloud features during training or evaluation, in contrast to task-specific and 3D LLM models that depend heavily on explicit 3D representations. In addition, our approach requires substantially fewer egocentric keyframes on average (2 compared to 8 in GPT4Scene~\cite{qi2025gpt4scene}), resulting in a more efficient and scalable training process. Compared to models designed for narrow tasks, our framework is more general and supports a wider range of spatial reasoning types, including relative direction and route planning, which are not covered by these benchmarks. It is also worth noting that the current evaluation metrics are rule-based and limited in expressiveness, which may not fully reflect a model’s capability in spatial understanding.

\begin{figure}
    \centering
    \includegraphics[width=\linewidth]{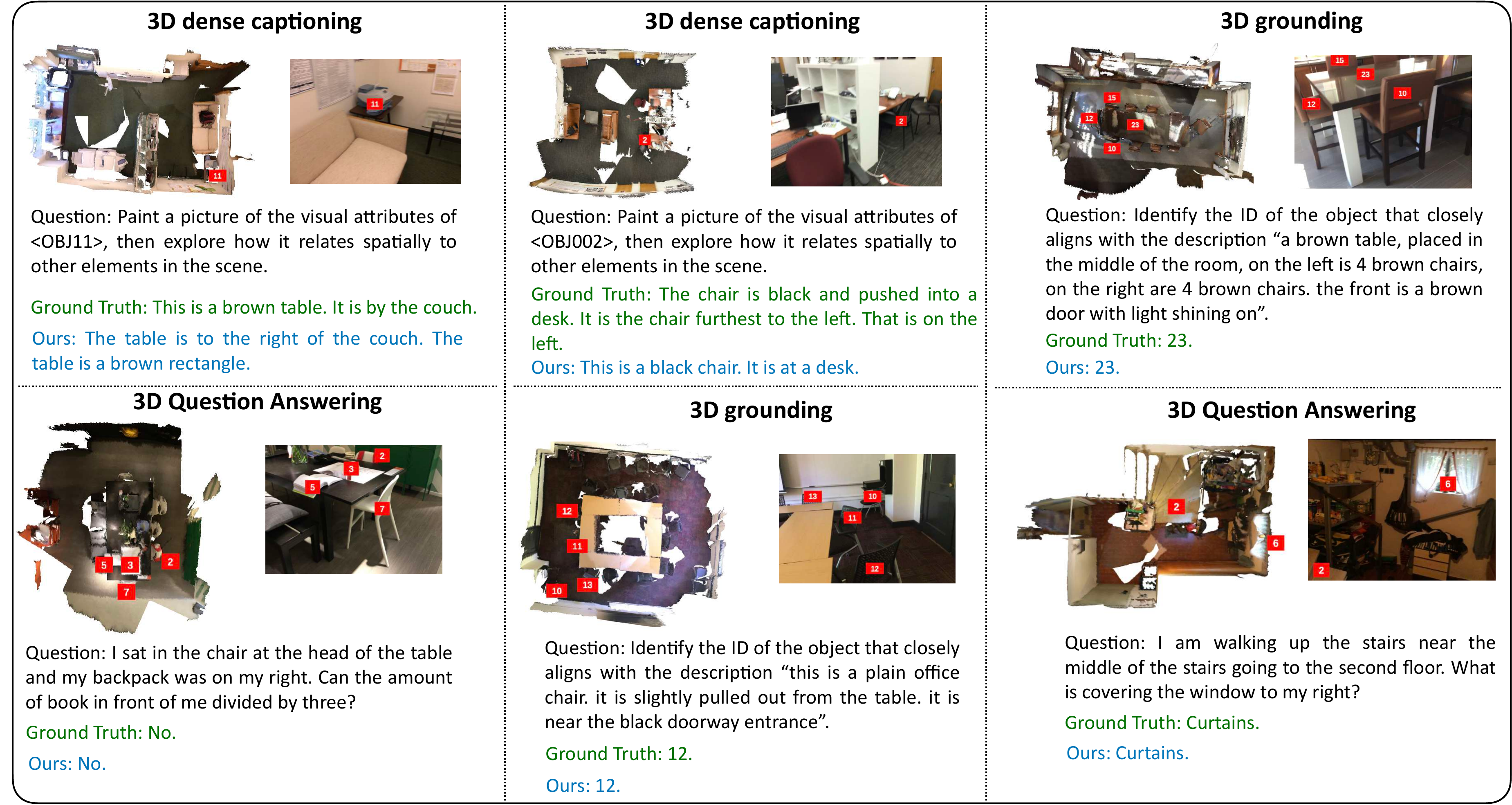}
    \caption{\label{fig:vis}\textbf{Output examples from our fine-tuned Qwen2.5-VL-7B model across multiple 3D spatial reasoning tasks.}The figure showcases model responses on 3D dense captioning, object grounding, and 3D question answering tasks. Each example includes the question, BEV and keyframe inputs with object marks, the ground-truth answer, and our model’s prediction. These examples illustrate the model’s ability to localize, describe, and reason about spatial relations using structured 2D prompts derived from 3D scenes. Across tasks, the model demonstrates strong alignment with ground-truth answers, even when questions require appearance attributes, relative spatial context, or numerical reasoning.}
    
\end{figure}

\noindent \textbf{Qualitative results.}
Figure~\ref{fig:vis} illustrates qualitative examples of our fine-tuned Qwen2.5-VL-7B model across three major spatial reasoning tasks: 3D dense captioning, object grounding, and 3D question answering. In each case, the model receives a BEV image with object marks, optionally supplemented with egocentric keyframes and metadata, and produces either a descriptive caption, an object ID, or a short-form answer. The examples demonstrate the model's ability to reason about visual attributes (\eg, ``a brown rectangle''), relative spatial positions (\eg, ``the table is to the right of the couch''), and numerical or commonsense questions. We observe that the model often produces answers consistent with the ground truth, and in some cases offers additional descriptive clarity grounded in the visual context. These results highlight the effectiveness of our \model{} prompting strategy in enabling rich spatial understanding from structured 2D inputs.

\section{Failure cases}

\begin{figure}
    \centering
    \includegraphics[width=0.9\linewidth]{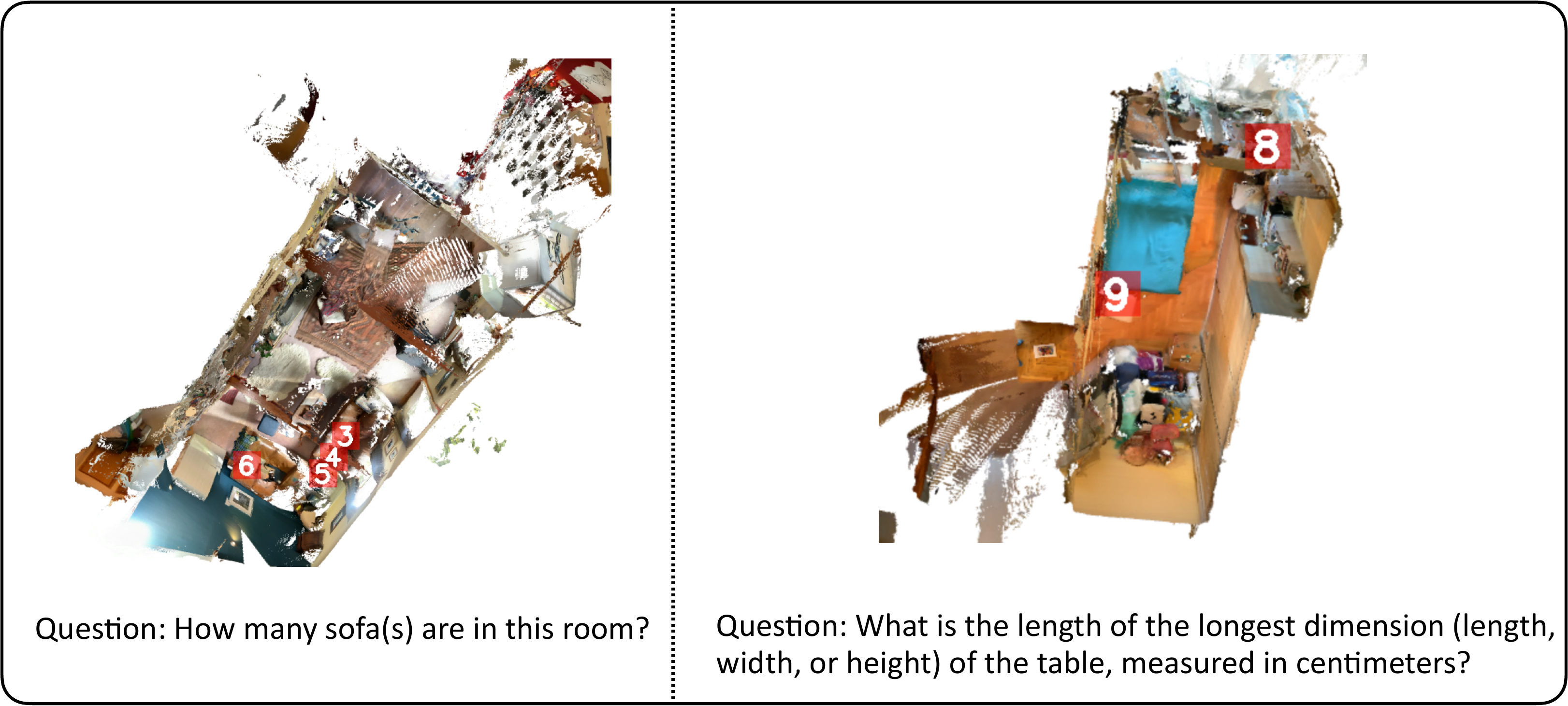}
    \caption{Examples of failure cases caused by 3D reconstruction and detection.}
    \label{fig:fail}
\end{figure}
To better understand the limitations of our approach, we conducted a qualitative error analysis on 30 representative questions spanning multiple QA types in VSI-Bench. Among the 16 failure cases, we identified two dominant causes. First, in 11 cases, the underlying 3D reconstruction was noisy or incomplete, producing degraded BEV projections that obscured critical spatial layouts. Second, in 5 cases, missing detections—often involving small or heavily occluded objects—led to incomplete structured inputs. Both factors reduce Struct2D’s ability to encode accurate spatial cues, thereby hindering its reasoning capability. Representative visualizations of these failure modes are shown in Figure~\ref{fig:fail}. These examples highlight the importance of robust 3D reconstruction and reliable object grounding for spatial reasoning in complex indoor scenes.

\section{Broader impacts}

Our work introduces \model{}, a perception-guided prompting framework that enables Multimodal Large Language Models (MLLMs) to perform robust spatial reasoning in 3D environments using only structured 2D and text inputs. This direction offers several broader implications for research, society, and the reasonable development of AI systems.

\textbf{Social Benefits.} \model{} lowers the barrier to 3D spatial reasoning by leveraging RGB-D perception instead of requiring dense 3D annotations or point cloud inputs during inference. This makes spatial understanding more accessible to a wide range of applications, especially in settings where real-time 3D sensing is noisy, sparse, or unavailable. Potential downstream applications include:
\begin{itemize}[nosep,leftmargin=2em]
    \item \textbf{Assistive robotics}, where spatial-language understanding is critical for navigation and object manipulation in dynamic home environments;
    \item \textbf{Augmented reality interfaces}, where natural-language spatial queries must be resolved in partially reconstructed environments;
    \item \textbf{Accessibility technologies}, especially for users with visual impairments, by enabling robust, language-driven scene understanding with minimal hardware.
\end{itemize}

\textbf{Potential Negative Impacts.} The preprocessing pipeline relies on egocentric video and 3D reconstruction, which may involve scenes from private homes or workplaces. If deployed in real-world applications, such systems may inadvertently capture sensitive spatial or personal data. Ensuring strict anonymization, access control, and user consent mechanisms is essential.

\textbf{Research Contributions.} By decoupling MLLM training from explicit 3D input requirements, \model{} promotes research into modular, scalable instruction-tuning pipelines that can generalize across environments with different sensor setups. Furthermore, our public release of \data{}---a large-scale spatial instruction dataset built with a principled blend of structured prompts, egocentric frames, and metadata---contributes valuable benchmarks to the broader vision-language community.

\end{document}